\documentclass[11pt]{article}

\PassOptionsToPackage{hyperfootnotes=false}{hyperref}
\usepackage[preprint]{acl}
\usepackage{times}
\usepackage{latexsym}
\usepackage[T1]{fontenc}
\usepackage[utf8]{inputenc}
\usepackage{microtype}
\usepackage{inconsolata}
\usepackage{graphicx}
\usepackage{booktabs}
\usepackage{amsmath}
\usepackage{tabularx}
\usepackage{colortbl}
\usepackage{threeparttable}
\usepackage{adjustbox}
\usepackage{float}
\usepackage{afterpage}
\usepackage[most]{tcolorbox}
\definecolor{oursgray}{gray}{0.94}
\definecolor{tablemainhead}{RGB}{232,240,250}
\definecolor{tablemainours}{RGB}{224,238,252}
\definecolor{tableanalysisrow}{RGB}{235,238,242}
\definecolor{tablerule}{RGB}{70,70,70}
\definecolor{RoyalBlue}{RGB}{65,105,225}

\newcommand{\wpgcell}[2]{\textbf{#1}{\tiny\,(#2)}}
\newcommand{\tablerowshade}{\smash{\makebox[0pt][l]{\textcolor{tableanalysisrow}{\rule[-0.45ex]{\linewidth}{2.2ex}}}}}

\title{WPG-MoE: Weak-Prior-Guided Dense Mixture-of-Experts for User-Level Social Media Depression Detection}

\author{
Xian Li$^{1,2}$ \quad
Yuanhe Tian$^{2}$\thanks{Corresponding Author.} \quad
Yang Yang$^{1}$ \quad
Guoqing Wang$^{1}$ \quad
Yan Song$^{3}$ \\
$^{1}$University of Electronic Science and Technology of China \\
$^{2}$Zhongguancun Academy \\
$^{3}$University of Science and Technology of China \\
\texttt{xianli@stu.uestc.edu.cn} \quad
\texttt{yhtian94@gmail.com} \\
\texttt{yang.yang@uestc.edu.cn} \quad
\texttt{gqwang0420@uestc.edu.cn} \quad
\texttt{clksong@gmail.com}
}

\begin{document}
\maketitle
\flushbottom

\begin{abstract}
Online social media posts provide scalable signals for early depression screening, and recent studies mainly improve pre-classification evidence through risk-post selection, symptom grounding, and clinically informed feature construction. However, these screening-stage designs often leave final decisions to a single detector, overlooking how users heterogeneously express depressive risk after screening. A monolithic classifier must average across heterogeneous users, which may dilute localized evidence and cause misclassification, especially for non-self-disclosing users. To address this issue, we propose WPG-MoE, a weak-prior-guided dense mixture-of-experts framework built on a shared large language model (LLM) backbone. WPG-MoE derives user-level weak semantic priors to softly route users to experts matched to different evidence layouts. We formulate this process as learning using privileged information (LUPI): rich LLM-extracted structured evidence guides training-time routing, while inference retains only Patient Health Questionnaire-9 (PHQ-9) template screening and the deployable backbone. Experiments on Chinese and English datasets show that WPG-MoE outperforms strong baselines with interpretable routing behavior. 
\end{abstract}

\section{Introduction}
\label{sec:introduction}

Depression affects an estimated 332 million people worldwide, yet treatment
coverage and minimally adequate care remain limited
\citep{moitra2022global,who-2025-depression}. Social media histories therefore
provide scalable early-identification signals, as \citet{de2013predicting}
showed for depression detection from naturalistic user traces
\citep{yates2017depression,shen2017depression,guntuku2017detecting,song2018feature,chancellor2020methods,nie2020named,diao2023hashtag,hu2024resemo}.

\begin{figure}[t]
    \centering
    \includegraphics[width=\linewidth]{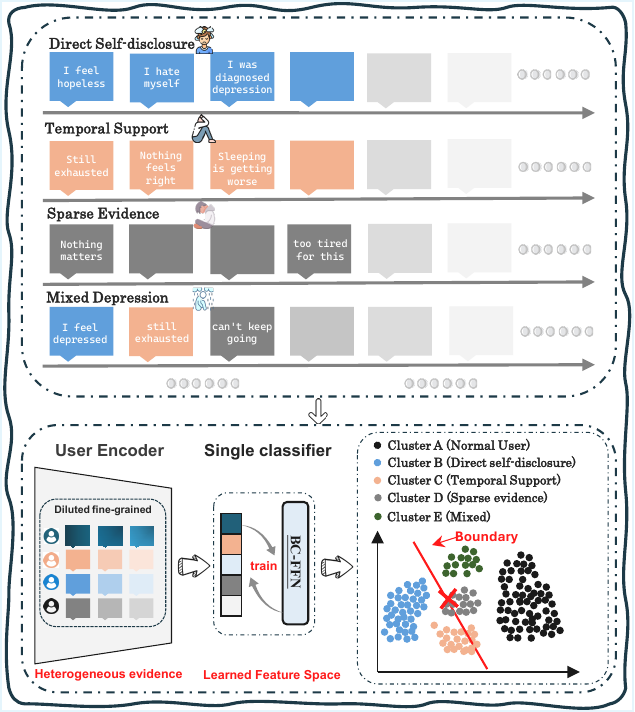}
    \caption{Heterogeneous evidence patterns create a mismatch for monolithic
    modeling.}
    \label{fig:intro-heterogeneity}
\end{figure}

Recent work improves user-level social media depression detection with
complete-history modeling and clinically structured evidence: multimodal fusion,
user-post summarization, symptom-aware temporal modeling, capsule-style
aggregation
\citep{shen2017depression,gui2019cooperative,zogan2021depressionnet,wang2022online,cai2023depression,liu2024depression},
Patient Health Questionnaire-9 (PHQ-9) or psychiatric-scale guidance
\citep{nguyen2022improving,zhang2022psychiatric,wang2025end}, and
large language model (LLM)-based annotation, summarization, retrieval, or
explanation for clinical evidence
\citep{wang2024explainable,lan2025depression,tian2024chimed,ravenda2025llms}. Yet final
predictors across pretrained language models (PLMs), sentence-embedding
pipelines, capsule models, tree classifiers, and retrieval-augmented LLM agents
still make one decision after screening.

The bottleneck is post-screening heterogeneity, not evidence
retrieval. Depressed users reveal risk through overlapping structures: diagnoses
or medication, sustained symptoms, or a few high-intensity posts amid otherwise
irrelevant histories
\citep{mendes2024identifying}. These are evidence structures, not hard clinical
subtypes. Figure \ref{fig:intro-heterogeneity} illustrates the
mismatch: after screening, a flat detector compresses heterogeneous signals into
one representation and boundary. This averaging can dilute localized evidence
and obscure weaker non-self-disclosure patterns, motivating dense conditional
specialization with soft routing, not hard partitioning.

Mixture-of-experts (MoE) naturally supports this specialization, but
task-loss-driven routing can be fragile in noisy mental-health settings. It
can favor generic correlations over clinically meaningful evidence patterns
\citep{shazeer2017outrageously,ma2018modeling,santos2023mental,dos2025mixture}.
We guide routing with weak priors from clinical cues,
encouraging specialization around self-disclosure, episode-supported evidence,
and sparse high-risk evidence.

We instantiate this as \textbf{WPG-MoE}, a \textbf{W}eak-\textbf{P}rior-
\textbf{G}uided dense \textbf{M}ixture-\textbf{o}f-\textbf{E}xperts framework
for user-level social media depression detection. WPG-MoE softly routes
users to evidence-matched experts over a shared LLM backbone. Because the
strongest priors come from external LLM-extracted structure useful in
training but costly at deployment, we cast WPG-MoE as learning using privileged
information (LUPI)
\citep{vapnik2009new}: privileged evidence constructs training-time priors and
evidence blocks, while inference keeps only deployable PHQ-9 screening, the
shared backbone, and history-level signals. Experiments on Chinese and English
datasets show improved prediction and interpretable routing. In summary, our
contributions follow:
\begin{itemize}
    \item We identify post-screening evidence heterogeneity as a failure mode in
    user-level social media depression detection.
    \item We introduce WPG-MoE, which softly routes users to experts with
    clinically grounded weak priors instead of hard subtype labels.
    \item We cast weak-prior-guided routing as LUPI: LLM-derived structure is
    training-only, while inference stays deployable through shared backbone, not
    external annotation pipelines.
\end{itemize}

\begin{figure*}[t]
\centering
\includegraphics[width=\textwidth]{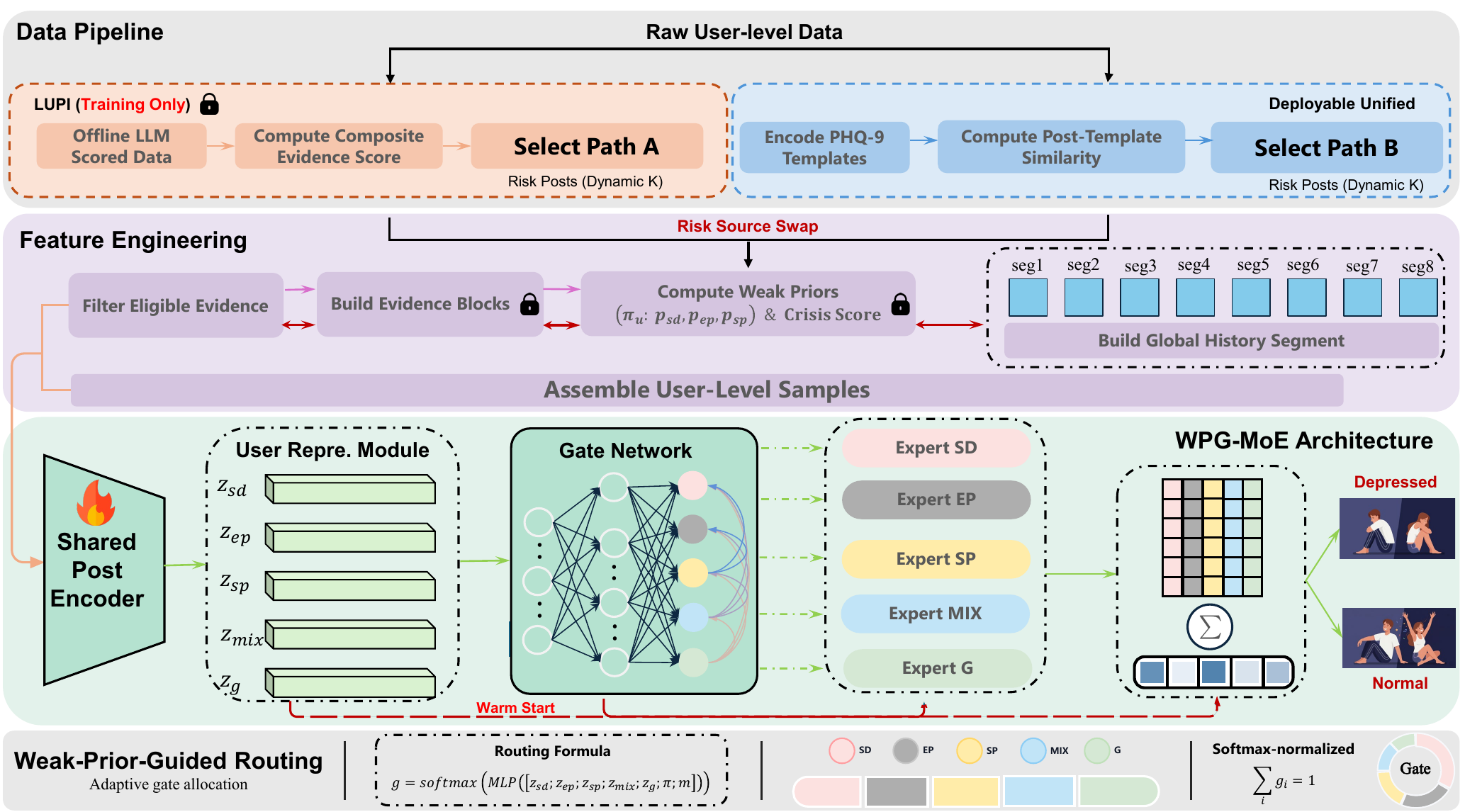}
\caption{Overall pipeline of WPG-MoE. Path A provides training-time privileged
evidence from an external scorer, Path B provides deployable PHQ-9 screening,
and inference keeps only Path B together with the shared backbone.}
\label{fig:wpg-moe-pipeline}
\end{figure*}

\section{The Approach}
\label{sec:approach}

\subsection{Problem Definition}

We study user-level textual depression detection. Given a user's
chronological posts $P=\{p_1,\ldots,p_n\}$ and binary label
$y \in \{0,1\}$, the task is to learn a predictor outputting a user-level
decision $\widehat{y}$:
\begin{equation}
\widehat{y} = \mathcal{F}(P).
\end{equation}
WPG-MoE uses privileged training and deployable inference: structured evidence
supervises training and routing; inference uses deployable textual signals.
Figure \ref{fig:wpg-moe-pipeline} summarizes this pipeline.

\subsection{Dual-Path Evidence Construction}

Construction follows the training--deployment gap: privileged
structure guides learning; PHQ-9 screening remains the inference interface.

We use two text-only risk-post paths: training-only Path A from offline
evidence scores and deployable Path B from PHQ-9 screening.
Both produce candidate set $R$, combined with history $H$
for evidence-centric prediction $\widehat{y}=f(R,H)$. Privileged training uses
$R^{A}$, whereas deployable inference uses $R^{B}$. Figure
\ref{fig:wpg-moe-pipeline} links both to prior induction, routing, and
prediction.

\paragraph{Path A:} Path A uses training-only structured signals
from an offline LLM \citep{alhamed2024using}. The Qwen3-max API scores
909,794 posts from SWDD
\citep{cai2023depression}, Twitter
\citep{shen2017depression}, and eRisk25
\citep{parapar2025erisk}. Each post receives symptom, crisis, anchor, duration,
confidence, and self-evidence attributes, then privileged post score $c_i$:
\begin{equation}
\begin{aligned}
c_i = \mathrm{Score}_A\big(&\phi_{\mathrm{sym}}(p_i),
\phi_{\mathrm{crisis}}(p_i),
\phi_{\mathrm{anchor}}(p_i), \\
&\phi_{\mathrm{dur}}(p_i),
\phi_{\mathrm{conf}}(p_i),
\phi_{\mathrm{self}}(p_i)\big).
\end{aligned}
\end{equation}
The $\phi$ terms denote six LLM-derived attributes.
$\mathrm{Score}_A(\cdot)$ maps them to
a $[0,1]$ scalar from symptom strength/coverage, crisis severity, anchor
presence, duration support, confidence, and explicit self-evidence. Ranking by
$c_i$ yields $R^{A}$ for auxiliary supervision/weak priors; the
external scorer remains separate from the deployable backbone. A blinded
human audit validates these fields: A/B agreement is substantial
($\kappa=0.671$--$0.709$), Qwen-to-adjudicated F1 exceeds 0.80 on main
fields, and user-level layout reaches $\kappa=0.654$ against
adjudicated labels (Appendix~\ref{sec:appendix-weak-prior-audit}).

\paragraph{Path B:} At inference, posts are scored against symptom templates.
Let $d \in \{1,\ldots,D\}$ index the $D$ PHQ-9 symptom dimensions, and
$\mathcal{T}_d$ be the template set for dimension $d$; Appendix
\ref{sec:appendix-phq9-templates} lists the templates. The
template screener computes
\begin{equation}
\begin{aligned}
s_{i,d} &= \max_{q \in \mathcal{T}_d}
\cos(\mathrm{Enc}_s(p_i), \mathrm{Enc}_s(q)), \\
r_i &= \mathrm{Score}_B(\{s_{i,d}\}_{d=1}^{D}),
\end{aligned}
\end{equation}
where $s_{i,d}$ is the maximum similarity between $p_i$ and dimension
$d$'s templates, and $\mathrm{Score}_B(\cdot)$ pools dimension-wise similarities
into deployable risk score $r_i$.
Candidates are selected dynamically:
\begin{equation}
\begin{aligned}
K &= \mathrm{Budget}(|P|), \\
R^{B} &= \mathrm{TopK}_{K}(\{r_i\}_{i=1}^{n}).
\end{aligned}
\end{equation}
$K$ is the candidate budget, and $\mathrm{Budget}(\cdot)$ is
history-length-aware. Following E2-LPS \citep{wang2025end}, the nominal
evidence-post screening ratio is 12.5\%, with a floor for short histories. The
shared backbone encodes candidates for user-level routing and prediction.

During training, routing and evidence scoring see Path-B signals alongside
privileged Path A. Table \ref{tab:alignment-perturbations} summarizes
four perturbations.

\begin{table}[t]
\centering
\scriptsize
\setlength{\tabcolsep}{3pt}
\renewcommand{\arraystretch}{1.0}
\begin{tabularx}{\linewidth}{@{}l c >{\raggedright\arraybackslash}X@{}}
\toprule
\textbf{Mechanism} & \textbf{Rate} & \textbf{Meaning / role} \\
\midrule
Risk Source Swap & 0.5 & Swaps Path-A and Path-B candidate sources so training also sees deployable risk-post quality. \\
META Dropout & 0.5 & Masks privileged \texttt{[META]} cues so post encoding does not depend on LLM-side annotations. \\
Episode Block Dropout & 0.4 & Removes privileged evidence blocks so the episode view can fall back to deployable risk-post evidence. \\
Prior Dropout & 0.3 & Zeros weak priors so routing remains usable when prior cues are absent at test time. \\
\bottomrule
\end{tabularx}
\caption{Training-time alignment mechanisms and rates.}
\label{tab:alignment-perturbations}
\end{table}

\subsection{Weak-Prior-Guided User Modeling}

Weak-prior modeling links screened evidence to conditional specialization,
mapping privileged layouts to routing tendencies and views.

\paragraph{Weak priors from coarse evidence tendencies and blocks.}
Post-screening heterogeneity has three soft layouts:
explicit self-disclosure, episode-supported evidence, and sparse high-risk
evidence. Explicit self-disclosure, common in mental-health self-reports, is modeled separately
\citep{de2014mental,coppersmith2015adhd,andalibi2017sensitive,harrigian2022then}. Episode-supported
evidence follows PHQ-9-style screening, but social media histories are too
sparse/irregular for rigid two-week rules
\citep{kroenke2001phq,macavaney2018rsdd,chen2018mood,trotzek2018utilizing,alhamed2024classifying,agarwal2025redepress};
we keep eligible first-person evidence posts above a composite-score
threshold and merge adjacent ones into temporal blocks $B$. Sparse
high-risk evidence captures users with only a few isolated but intense signals,
observed in post-level monitoring and symptom-detection work
\citep{jamil2017monitoring,yadav2020identifying,jiang2020detection}. From
privileged posts $R^{A}$ and blocks $B$, we derive
$\pi=[\pi_{\mathrm{self}}, \pi_{\mathrm{epis}}, \pi_{\mathrm{sparse}}]$, whose
entries summarize self-disclosure, episode-supported evidence, and sparse
high-risk evidence:
\begin{equation}
\pi = [\pi_{\mathrm{self}},\, \pi_{\mathrm{epis}},\, \pi_{\mathrm{sparse}}]
= \mathrm{Prior}(R^{A}, B).
\end{equation}
$\mathrm{Prior}(\cdot)$ scores self-disclosure from self-claims, anchors,
and confidence in $R^{A}$; episode support from the strongest block in $B$; and
sparse evidence when neither tendency dominates and few high-scoring posts
remain. The audited cues support soft routing, not hard labels.\footnote{The
user-level audit shows substantial agreement and stable
weak-prior-to-reference alignment; Appendix
\ref{sec:appendix-user-type-audit} reports the statistics.}
In parallel, history is partitioned into eight chronological segments
$\{P^{(j)}\}_{j=1}^{8}$. Segment encoder $\mathrm{SegEnc}(\cdot)$ mean-pools
shared post encodings into $u^{(j)}$; the temporal aggregator
$\mathrm{TAgg}(\cdot)$ applies self-attention over the eight segment vectors to
produce global history representation $H$:
\begin{equation}
\begin{aligned}
u^{(j)} &= \mathrm{SegEnc}(P^{(j)}), \qquad j=1,\ldots,8, \\
H &= \mathrm{TAgg}\!\big(u^{(1)}, \ldots, u^{(8)}\big),
\end{aligned}
\end{equation}
This branch retains broad mood evolution more cheaply than encoding all posts
\citep{cai2023depression,agarwal2025redepress}.

\paragraph{Multi-view user representation.}
We use five views: three tendency-specific channels for self-disclosure,
episode-supported evidence, and sparse high-risk evidence; a mixed cross-channel
view; and a global residual-history view. The shared LLM encoder represents
each candidate post with its last non-padding token's final-layer hidden state.
The user module builds $\{z_1, z_2, z_3, z_4, z_5\}$: three
attention-pooled channel views, a mean-pooled mixed risk-post view, and a global
view combining temporal self-attention and projected summary statistics.
We collect crisis and history statistics into metadata vector $m$.
A five-expert dense MoE predicts gate weights
$g=[g_1,\ldots,g_5]$, where $e_k$ is the $k$-th expert output and
$h$ the fused state:
\begin{equation}
\begin{aligned}
g &= \mathrm{softmax}\Big(\mathrm{MLP}([z_1; z_2; z_3; z_4; z_5; \pi; m])\Big), \\
e_k &= \mathrm{Expert}_k([z_k; m]), \\
h &= \sum_{k=1}^{5} g_k\, e_k,
\end{aligned}
\end{equation}
Weak priors act as soft routing hints rather than hard subtype labels.

\subsection{Prediction and Learning}

Learning preserves the LUPI boundary: privileged evidence supervises training,
while deployable posts define the routed state used at inference.

\noindent\textbf{Evidence scoring.} Given candidate post representation
$x_i$, the fused user state $h$, and the gate vector $g$, the auxiliary
scoring head computes training-time post score:
\begin{equation}
a_i = \sigma\Big(\mathrm{MLP}([x_i; h; g])\Big).
\end{equation}
Here $a_i \in (0,1)$ supervises training-time evidence.

\noindent\textbf{Training objective.} We jointly optimize user prediction,
routing, and evidence supervision:
\begin{equation}
\mathcal{L} = \mathcal{L}_{\mathrm{cls}}
+ \alpha \mathcal{L}_{\mathrm{route}}
+ \beta \mathcal{L}_{\mathrm{evi}}
+ \gamma \mathcal{L}_{\mathrm{bal}}
+ \delta_t \mathcal{L}_{\mathrm{ent}}.
\end{equation}
Here $\delta_t$ is the training-step-dependent entropy coefficient.
$\mathcal{L}_{\mathrm{cls}}$ is the user-level classification loss,
$\mathcal{L}_{\mathrm{route}}$ aligns the first three expert gates with
confident weak priors, $\mathcal{L}_{\mathrm{evi}}$ supervises evidence scores
on depressed users with silver labels, $\mathcal{L}_{\mathrm{bal}}$ discourages
expert imbalance, and $\mathcal{L}_{\mathrm{ent}}$ controls routing sharpness
through decayed entropy.

\noindent\textbf{Inference.} Training applies risk-source swap/dropout to
metadata, evidence blocks, weak priors, and candidate posts, exposing gates to
privileged and deployable inputs. At inference, removing Path A/privileged
annotations leaves only template-screened posts, shared backbone,
eight-segment summaries, and
user statistics. Appendix~\ref{sec:appendix-implementation-details} reports
rules and hyperparameters for candidate construction, weak-prior
induction, routing, and training.

\section{Experimental Settings}
\label{sec:experimental-settings}

\subsection{Datasets, Label Audit, and Controlled Protocol}

We use three text-only user-level depression-detection datasets: SWDD (Chinese;
\citealp{cai2023depression}), the Twitter depression dataset (English;
\citealp{shen2017depression}), and eRisk25 (English;
\citealp{parapar2025erisk}). All runs use a unified stratified
$80/10/10$ holdout for within- and cross-dataset comparison, following recent
user-level depression-detection practice with consistent partitions
\citep{nguyen2022improving,liu2024depression,wang2025end,bucur2025datasets};
the protocol supports controlled, not leaderboard, comparison.

\paragraph{SWDD label audit.} SWDD's raw self-report flag
(\texttt{self\_reported}) contains label errors, so all SWDD experiments use
the corrected split unless otherwise noted. Candidate corrections from
mismatch screens were manually reviewed per user before finalizing corrected
counts. The audit procedure, corrected class distribution, and
representative cases appear in
Appendix~\ref{sec:appendix-swdd-noise-examples} and Table
\ref{tab:swdd-noise-audit}. Table \ref{tab:dataset-stats} summarizes the final
dataset versions.

\begin{table}[t]
\centering
\scriptsize
\setlength{\tabcolsep}{2pt}
\renewcommand{\arraystretch}{1.04}
\resizebox{\linewidth}{!}{
\begin{tabular}{@{}lccc@{}}
\toprule
\textbf{Statistic} & \textbf{SWDD} & \textbf{Twitter} & \textbf{eRisk25} \\
\midrule
No. of users (Dep.) & 3,711 & 1,218 & 102 \\
No. of users (Ctrl.) & 19,526 & 1,273 & 807 \\
No. of posts (Dep.) & 643,139 & 196,268 & 70,387 \\
No. of posts (Ctrl.) & 3,357,273 & 700,320 & 348,803 \\
Avg. posts per user (Dep.) & 173.3 & 161.1 & 690.1 \\
Avg. posts per user (Ctrl.) & 171.9 & 550.1 & 432.2 \\
Users split & 18,588 / 2,324 / 2,325 & 1,992 / 249 / 250 & 726 / 91 / 92 \\
\bottomrule
\end{tabular}
}
\caption{Statistics of the three datasets used in the final round.}
\label{tab:dataset-stats}
\end{table}

\begin{figure*}[!t]
\centering
\includegraphics[width=\textwidth]{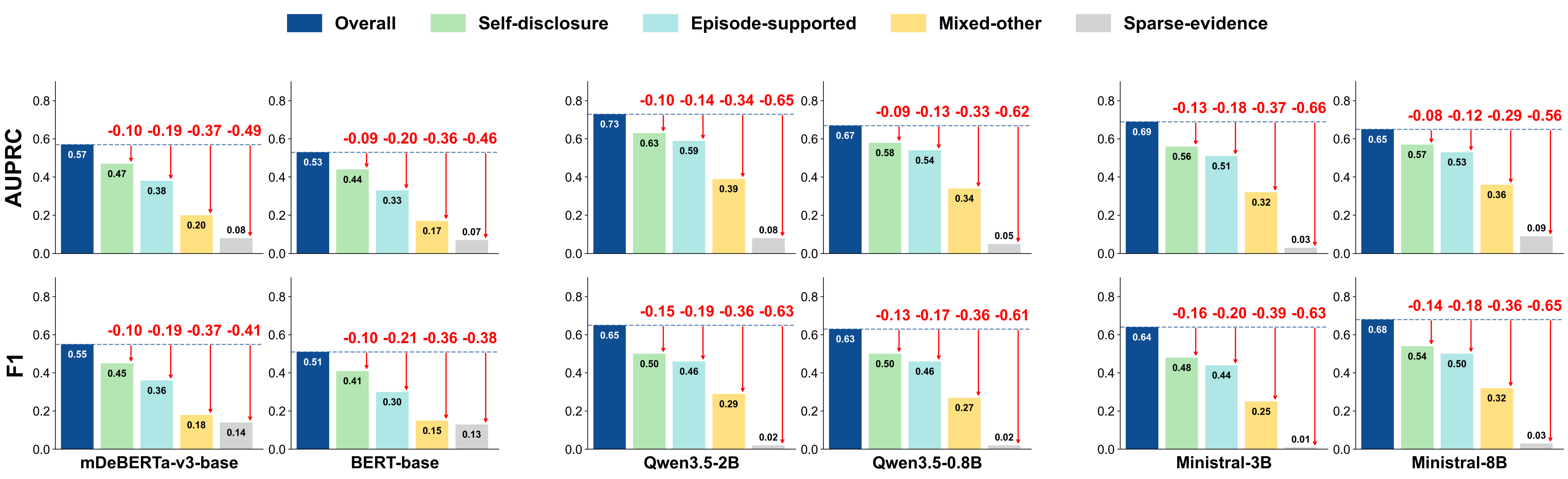}
\caption{Human-labeled evidence-layout diagnostic. Six flat detectors' AUPRC/F1 on a
blinded cross-dataset set: Overall uses the audit pool; type columns use
adjudicated \textit{self-disclosure},
\textit{episode-supported}, \textit{mixed/other}, and \textit{sparse-evidence}
positives against fixed controls.}
\label{fig:exp1-slice-auprc}
\end{figure*}

\begin{figure}[t]
\centering
\includegraphics[width=\linewidth]{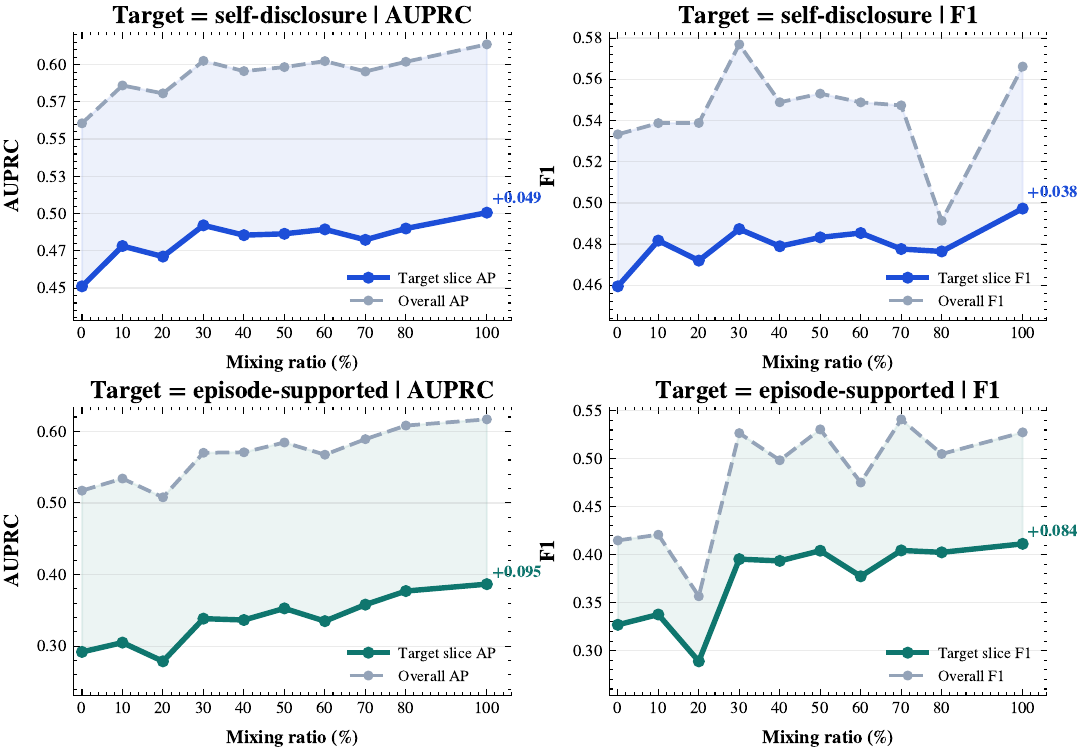}
\caption{Controlled mixing on SWDD. Top row targets \textit{self-disclosure}
and bottom row \textit{episode-supported}; each reports target-slice AUPRC/F1
as cross-slice positives are mixed into training.}
\label{fig:controlled-mixing-routing}
\vspace{-8pt}
\end{figure}

\subsection{Experimental Protocols}

All experiments use unified holdout splits in Table
\ref{tab:dataset-stats}. Models train on one source and test in-domain or
transfer without changes. We
report five-run mean recall, F1, area under the receiver operating
characteristic curve (AUROC), and area under the precision-recall curve (AUPRC).
Recall and F1 measure thresholded detection, AUROC measures
threshold-free ranking, and AUPRC targets imbalanced settings such as SWDD and
eRisk25.

We also test whether post-screening evidence layouts yield stable differences
without weak-prior-rule slice sources. The diagnostic uses blinded human
evidence-layout labels: 360 depressed users from SWDD, Twitter, and eRisk25
paired with 1,800 fixed same-dataset controls.
Because sparse/mixed layouts are rare within datasets, we pool
three datasets and omit per-dataset layout-prevalence estimates.
Appendix \ref{sec:appendix-diagnostic-splits} reports complementary
large-scale rule-derived SWDD diagnostic and controlled-mixing counts. For
backbone replacement, Base is a raw backbone classifier and Ours denotes
same-backbone WPG-MoE. We report
mDeBERTa-v3-base \citep{he2023debertav3}, Qwen3.5-2B \citep{qwen35}, and
Ministral-3B \citep{liu2026ministral3}; Appendix
\ref{sec:appendix-full-backbone-replacement} adds BERT-base
\citep{devlin-etal-2019-bert}, Qwen3.5-0.8B \citep{qwen35}, and Ministral-8B
\citep{liu2026ministral3}. Unless replaced, WPG-MoE uses Qwen3.5-2B as
shared encoder; Path A is separate and used only for source-train depressed
users, while dev/test and transfer targets use raw histories and Path B.

\subsection{Baselines and Metrics}

We compare WPG-MoE with seven baselines in four families:
PHQ-9-grounded pattern matching \citep{nguyen2022improving},
psychiatric-scale-guided risky-post screening
\citep{zhang2022psychiatric,wang2025end}, symptom-structured representation
learning \citep{liu2024depression}, and LLM-assisted clinically informed
detection \citep{lan2025depression}. Detailed baseline notes are given in
Appendix~\ref{sec:appendix-baselines}.

\vspace{-8pt}
\section{Results and Analyses}
\label{sec:results-analyses}

\subsection{Slice-Level Heterogeneity}

\begin{table}[t]
\centering
\footnotesize
\setlength{\tabcolsep}{2pt}
\renewcommand{\arraystretch}{1.04}
\begin{tabular*}{\linewidth}{@{\extracolsep{\fill}}lccccc@{}}
\toprule
\textbf{Scope} & \textbf{Users} & \textbf{Raw agr.} &
\textbf{$\kappa$} & \textbf{$\alpha$} & \textbf{Adj.} \\
\midrule
SWDD & 180 & 0.817 & 0.735 & 0.738 & 33 \\
Twitter & 120 & 0.792 & 0.690 & 0.694 & 25 \\
eRisk25 & 60 & 0.767 & 0.641 & 0.650 & 14 \\
\midrule
Overall & 360 & 0.800 & 0.707 & 0.708 & 72 \\
\bottomrule
\end{tabular*}
\caption{Human evidence-layout agreement. Adj. denotes A/B disagreements
adjudicated by Annotator C.}
\label{tab:human-layout-agreement}
\end{table}

To test whether screened data contains evidence heterogeneity affecting
detector performance before WPG-MoE, we evaluate six flat detectors from three
model families (two variants each) on a blinded human evidence-layout diagnostic.
Slices are assigned from raw, unscored packets under
pre-specified annotation guidelines, rather than rule-derived categories. The
diagnostic pools 360 depressed users from SWDD, Twitter, and eRisk25 with the
same 1,800 controls and validation-set thresholding per slice;
sparse/mixed cases are rare, motivating cross-dataset pooling. Table
\ref{tab:human-layout-agreement} reports raw agreement 0.800, Cohen's
$\kappa=0.707$, and Krippendorff's $\alpha=0.708$; Appendix
\ref{sec:appendix-diagnostic-splits} gives sampling, packet visibility, labels,
guidelines, and the larger corrected-SWDD tendency/block diagnostic. Figure
\ref{fig:exp1-slice-auprc} shows stable degradation across model families:
\textit{self-disclosure} and \textit{episode-supported} users are easier,
whereas \textit{mixed/other} and \textit{sparse-evidence} users are more
difficult. The same ordering in the larger diagnostic further indicates
evidence layout is tied to detection difficulty.

\subsection{Overall Results and Generalization}

We evaluate within-dataset detection and transfer under identical
partitions. Table
\ref{tab:cross-dataset-target-tuned} compares WPG-MoE with seven baselines under
the unified holdout for method comparability, not leaderboard
claims. WPG-MoE gives the strongest in-domain results and keeps advantages on
most cross-dataset Recall, F1, and AUPRC cells. Transfer gains are clearest when
SWDD is source,
consistent with its larger supervision scale. Twitter-trained models remain
competitive, while eRisk25-trained models leave a few off-diagonal AUROC cells
matched or slightly exceeded by clinically guided baselines, suggesting robust
transfer needs enough source supervision under shift.
Appendix~\ref{sec:appendix-seedwise-reliability} reports Ours across five seeds;
F1 and AUPRC standard deviations stay within 0.0063--0.0083 and
0.0088--0.0113 across the nine train$\rightarrow$test cells.

\begin{table*}[t]
\centering
\begin{threeparttable}
\small
\setlength{\tabcolsep}{3pt}
\renewcommand{\arraystretch}{1.06}
\begingroup
\arrayrulecolor{tablerule}
\begin{adjustbox}{width=\textwidth,center}
\begin{tabular}{@{}llcccc@{\hspace{8pt}}cccc@{\hspace{8pt}}cccc@{}}
\toprule
\rowcolor{tablemainhead}
\textbf{Train} & \textbf{Method} & \multicolumn{4}{c}{\textbf{Test on SWDD}} & \multicolumn{4}{c}{\textbf{Test on Twitter}} & \multicolumn{4}{c}{\textbf{Test on eRisk25}} \\
\cmidrule(lr){3-6}\cmidrule(lr){7-10}\cmidrule(lr){11-14}
\rowcolor{tablemainhead}
\textbf{on} &  & \textbf{Rec.} & \textbf{F1} & \textbf{AUROC} & \textbf{AUPRC} & \textbf{Rec.} & \textbf{F1} & \textbf{AUROC} & \textbf{AUPRC} & \textbf{Rec.} & \textbf{F1} & \textbf{AUROC} & \textbf{AUPRC} \\
\midrule
\textbf{SWDD} & Pattern (threshold) & 0.6388 & 0.6046 & 0.9270 & 0.6799 & 0.5410 & 0.4806 & 0.6117 & 0.5094 & 0.6275 & 0.6302 & 0.8170 & 0.5875 \\
 & Pattern (CNN) & 0.6565 & 0.6871 & 0.9373 & 0.8011 & 0.6950 & 0.5920 & 0.6877 & 0.5657 & 0.5590 & 0.5789 & 0.7824 & 0.5747 \\
 & HAN-BERT(Psych) & 0.5438 & 0.5318 & 0.8624 & 0.6261 & 0.6142 & 0.5410 & 0.6312 & 0.5465 & 0.6161 & 0.4531 & 0.7958 & 0.3503 \\
 & Bert(Clus+Abs) & \underline{0.7224} & \underline{0.7436} & \underline{0.9453} & \underline{0.8092} & \underline{0.7501} & \underline{0.6410} & \underline{0.7980} & \underline{0.6398} & \underline{0.7727} & 0.2848 & 0.5920 & 0.2150 \\
 & E2-LPS & 0.5099 & 0.4428 & 0.7820 & 0.4650 & 0.4480 & 0.3810 & 0.5788 & 0.4980 & 0.4888 & 0.3631 & 0.7395 & 0.3198 \\
 & DeCapsNet & 0.5903 & 0.6075 & 0.9078 & 0.7001 & 0.4920 & 0.4430 & 0.5580 & 0.5019 & 0.7457 & \underline{0.6850} & \underline{0.8410} & \underline{0.6028} \\
 & DORIS & 0.6080 & 0.5774 & 0.9041 & 0.7045 & 0.5810 & 0.5120 & 0.6535 & 0.5710 & 0.4610 & 0.4911 & 0.7712 & 0.4474 \\
\rowcolor{tablemainours}
 & \textbf{Ours} & \textbf{0.8050} & \textbf{0.7490} & \textbf{0.9490} & \textbf{0.8300} & \textbf{0.7860} & \textbf{0.6680} & \textbf{0.8270} & \textbf{0.6610} & \textbf{0.8040} & \textbf{0.7110} & \textbf{0.8830} & \textbf{0.6310} \\
\midrule
\textbf{Twitter} & Pattern (threshold) & 0.6687 & 0.5601 & 0.8013 & 0.6213 & 0.6110 & 0.5639 & 0.6677 & 0.5898 & 0.5975 & 0.5402 & 0.7909 & 0.5453 \\
 & Pattern (CNN) & 0.7287 & \underline{0.6341} & \textbf{0.8720} & \underline{0.7217} & 0.7250 & 0.6827 & 0.8347 & 0.7621 & 0.4210 & 0.4342 & 0.6780 & 0.4085 \\
 & HAN-BERT(Psych) & 0.6577 & 0.3002 & 0.5436 & 0.2496 & 0.7350 & 0.6542 & 0.8303 & 0.7243 & \underline{0.7831} & \underline{0.6549} & \underline{0.8761} & \underline{0.6177} \\
 & Bert(Clus+Abs) & \underline{0.7354} & 0.3864 & 0.6091 & 0.3475 & 0.6710 & 0.6151 & 0.7565 & 0.7008 & 0.6835 & 0.3578 & 0.6952 & 0.2780 \\
 & E2-LPS & 0.6949 & 0.3358 & 0.4612 & 0.2350 & 0.7410 & \underline{0.7410} & \underline{0.9039} & \underline{0.8206} & 0.5680 & 0.4896 & 0.7364 & 0.4550 \\
 & DeCapsNet & 0.7184 & 0.4370 & 0.4975 & 0.3458 & 0.5620 & 0.5136 & 0.6020 & 0.5050 & 0.6663 & 0.5931 & 0.8178 & 0.5774 \\
 & DORIS & 0.6263 & 0.4702 & 0.6155 & 0.3989 & \underline{0.7610} & 0.7115 & 0.8635 & 0.7755 & 0.4402 & 0.3955 & 0.7282 & 0.4394 \\
\rowcolor{tablemainours}
 & \textbf{Ours} & \textbf{0.7890} & \textbf{0.6530} & \underline{0.8520} & \textbf{0.7420} & \textbf{0.8440} & \textbf{0.7630} & \textbf{0.9290} & \textbf{0.8520} & \textbf{0.8140} & \textbf{0.6890} & \textbf{0.9090} & \textbf{0.6880} \\
\midrule
\textbf{eRisk25} & Pattern (threshold) & 0.6087 & 0.4901 & 0.7857 & 0.4931 & 0.4947 & 0.4206 & 0.5737 & 0.4543 & 0.6682 & 0.5926 & 0.7857 & 0.5430 \\
 & Pattern (CNN) & \textbf{0.7450} & \underline{0.5519} & \underline{0.8500} & \underline{0.4941} & 0.6592 & 0.5560 & 0.7044 & 0.5913 & \underline{0.7950} & \underline{0.7150} & \underline{0.8982} & \underline{0.6828} \\
 & HAN-BERT(Psych) & 0.6900 & 0.3255 & 0.5424 & 0.2228 & \underline{0.7512} & \underline{0.6472} & \underline{0.7964} & \underline{0.6844} & 0.7150 & 0.6386 & 0.8348 & 0.5860 \\
 & Bert(Clus+Abs) & 0.6784 & 0.4312 & 0.6894 & 0.3792 & 0.5475 & 0.4626 & 0.6113 & 0.5020 & 0.7550 & 0.6750 & 0.8656 & 0.6414 \\
 & E2-LPS & 0.6344 & 0.3828 & 0.6541 & 0.3167 & 0.7029 & 0.6099 & 0.7656 & 0.6642 & 0.5550 & 0.4933 & 0.6846 & 0.4093 \\
 & DeCapsNet & 0.6711 & 0.2857 & 0.5350 & 0.2050 & 0.4300 & 0.3668 & 0.5180 & 0.4050 & 0.4773 & 0.4136 & 0.6350 & 0.3150 \\
 & DORIS & 0.5442 & 0.3562 & 0.6200 & 0.2925 & 0.6007 & 0.5093 & 0.6712 & 0.5666 & 0.6191 & 0.5400 & 0.7406 & 0.4842 \\
\rowcolor{tablemainours}
 & \textbf{Ours} & \underline{0.7340} & \textbf{0.5690} & \textbf{0.8550} & \textbf{0.6180} & \textbf{0.7590} & \textbf{0.6580} & \textbf{0.8120} & \textbf{0.7020} & \textbf{0.8210} & \textbf{0.7440} & \textbf{0.9260} & \textbf{0.7350} \\
\bottomrule
\end{tabular}
\end{adjustbox}
\endgroup
\caption{Cross-dataset comparison of WPG-MoE and seven baselines under the
controlled unified holdout protocol. Bold marks the best value within each
train block, test dataset, and metric column; underlining marks the corresponding
second-best value.}
\label{tab:cross-dataset-target-tuned}
\end{threeparttable}
\end{table*}

\begin{figure*}[!t]
\centering
\begin{minipage}[t]{0.48\textwidth}
\centering
\includegraphics[width=0.94\linewidth]{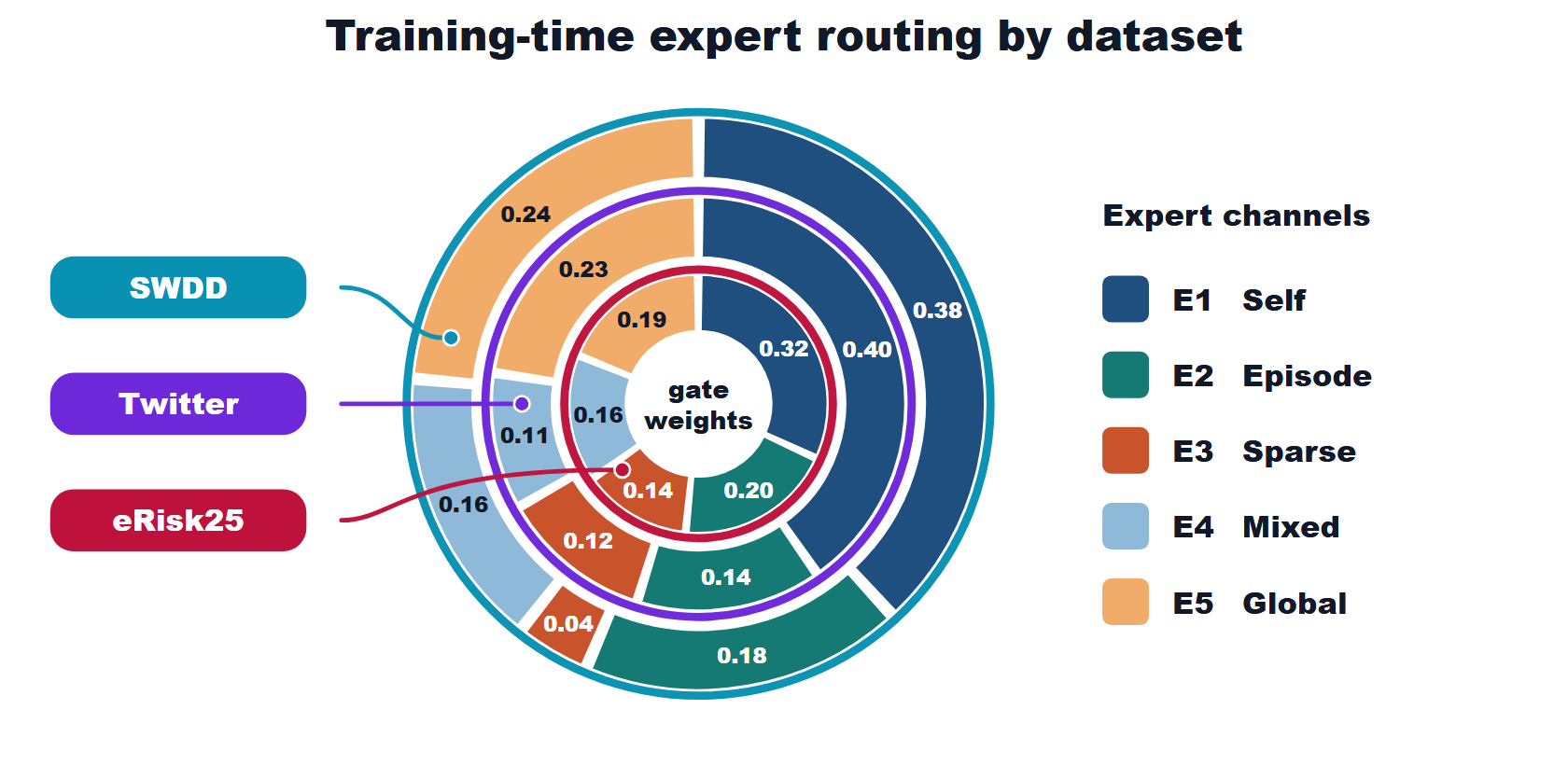}
\caption{Average training-time gate weights. E1--E5 denote the self-disclosure,
episode-supported, sparse-evidence, mixed, and global experts.}
\label{fig:gate-weights-nested-donut}
\end{minipage}\hfill
\begin{minipage}[t]{0.48\textwidth}
\centering
\includegraphics[width=0.88\linewidth]{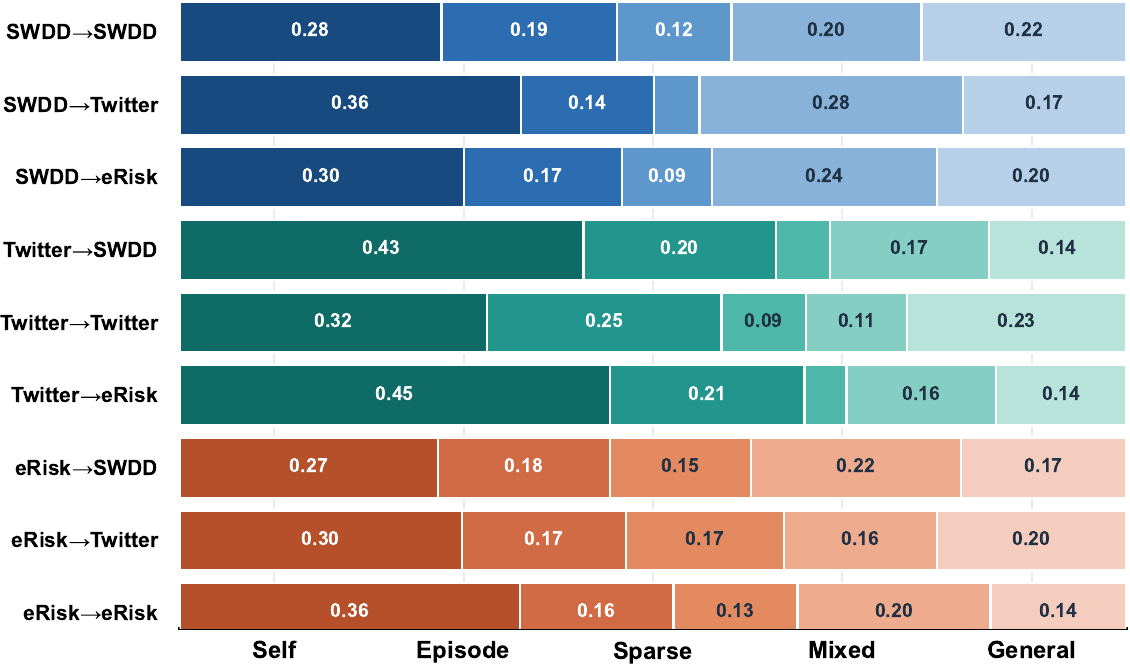}
\caption{Average test-time gate allocations across train$\rightarrow$test
settings.}
\label{fig:test-gate-allocations}
\end{minipage}
\end{figure*}

These results use Qwen3.5-2B as the default deployable backbone. To
separate architecture effects from backbone choice,
Table \ref{tab:backbone-replacement} compares each Base classifier with
matched-backbone WPG-MoE on in-domain cells; Appendix
\ref{sec:appendix-full-backbone-replacement} gives the complete
train$\rightarrow$test matrix and compact-LLM variants. Ours
improves over the corresponding Base setting for mDeBERTa-v3-base, Qwen3.5-2B,
and Ministral-3B, tying the gain to WPG-MoE rather than selecting one
stronger pretrained model.

\subsection{Controlled Mixing and Routing Analyses}

Routing analysis tests whether layouts act as compatible tendencies,
not isolated classes, via slice mixing and train$\rightarrow$test gate
allocation.
 
\paragraph{Controlled mixing and dense routing.}
Slice analysis shows stable layout differences, but layouts need not be
isolated classes. To test shared depressive signal, we start from one target
slice and add positives from other slices during training. Figure
\ref{fig:controlled-mixing-routing} shows target-slice AUPRC and F1 usually hold
or improve, most clearly when \textit{self-disclosure} users enter
\textit{episode-supported} training. This supports dense MoE: overlapping
evidence tendencies enable expert-signal sharing while retaining
tendency-aware specialization.

\paragraph{Gate allocation analysis.}
Figure \ref{fig:gate-weights-nested-donut} reports average training gates
for five WPG-MoE expert views on SWDD, Twitter, and eRisk25:
self-disclosure receives the largest weight, reflecting strong weak-prior
self-report cues, while episode, mixed, and global experts remain active,
avoiding reliance only on explicit diagnosis or medication mentions. Figure
\ref{fig:test-gate-allocations} reports test-time allocations;
train$\rightarrow$test variation suggests gates adapt evidence mixtures to
target users rather than follow fixed patterns. Dense-MoE weights describe
expert-view contributions, not evidence-type assignments.


\subsection{Ablation Analysis}
\label{sec:ablation}
\vspace{0.5\baselineskip}

\begin{figure*}[!t]
\centering
\captionsetup{hypcap=false}
\includegraphics[width=\textwidth]{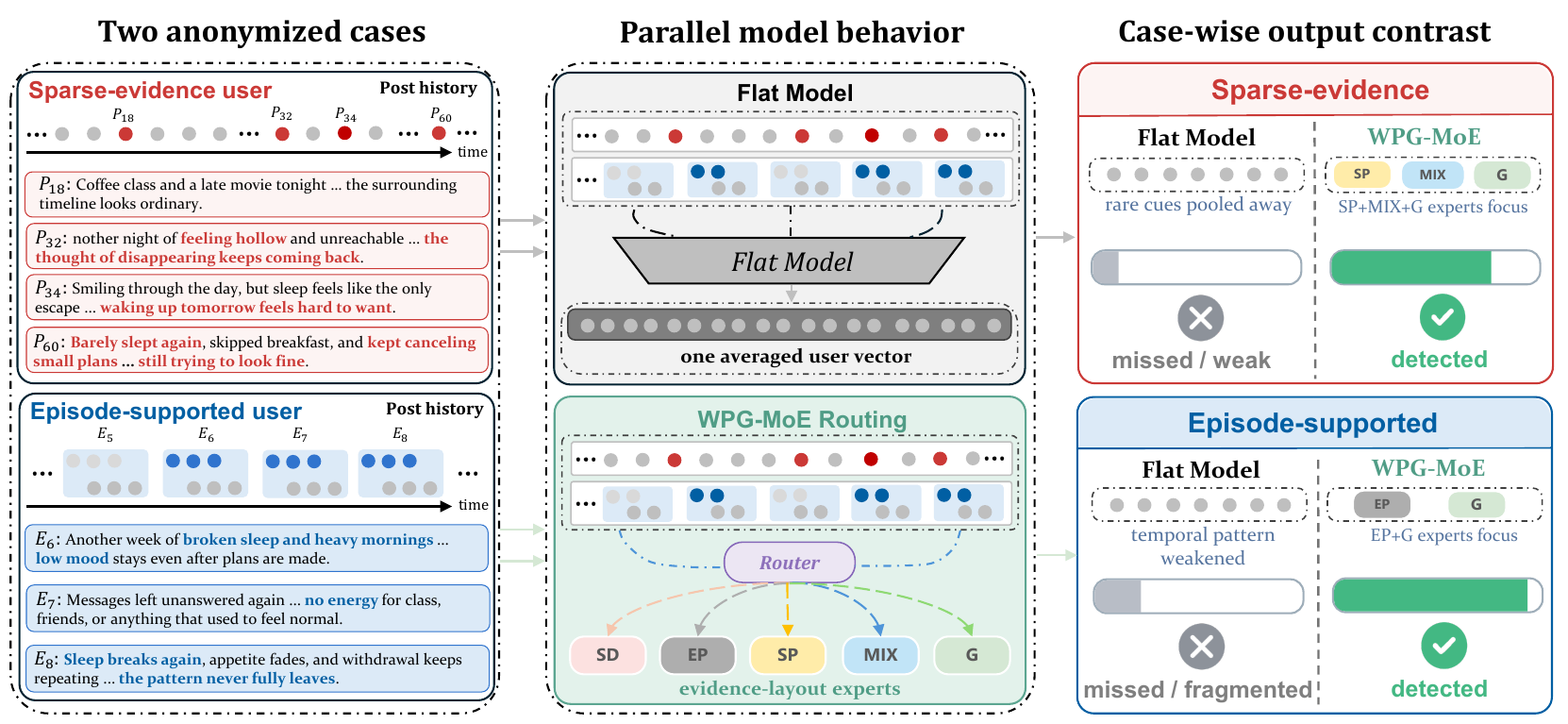}
\captionof{figure}{Case-wise comparison on two anonymized SWDD users. The flat model
denotes DeCapsNet; SP, EP, MIX, and G denote sparse-evidence,
episode-supported, mixed, and global experts.}
\label{fig:case-study}
\vspace{0.75\baselineskip}

\centering
\begin{minipage}[t]{\columnwidth}
\centering
\begin{threeparttable}
\scriptsize
\setlength{\tabcolsep}{1.2pt}
\renewcommand{\arraystretch}{1.02}
\begingroup
\arrayrulecolor{tablerule}
\begin{adjustbox}{width=\linewidth,center}
\begin{tabular}{@{}llcccc@{}}
\toprule
\textbf{Data} & \textbf{Method} & \textbf{Rec.} & \textbf{F1} & \textbf{AUROC} & \textbf{AUPRC} \\
\midrule
\textbf{SWDD} & mDeBERTa-v3-base (Base) & 0.64 & 0.58 & 0.73 & 0.60 \\
\rowcolor{tableanalysisrow}
 & mDeBERTa-v3-base (\textbf{Ours}) & \wpgcell{0.76}{+0.12} & \wpgcell{0.71}{+0.13} & \wpgcell{0.86}{+0.13} & \wpgcell{0.74}{+0.14} \\
 & Qwen3.5-2B (Base) & 0.74 & 0.67 & 0.87 & 0.75 \\
\rowcolor{tableanalysisrow}
 & Qwen3.5-2B (\textbf{Ours}) & \wpgcell{0.81}{+0.07} & \wpgcell{0.75}{+0.07} & \wpgcell{0.95}{+0.08} & \wpgcell{0.83}{+0.08} \\
 & Ministral-3B (Base) & 0.71 & 0.67 & 0.83 & 0.72 \\
\rowcolor{tableanalysisrow}
 & Ministral-3B (\textbf{Ours}) & \wpgcell{0.79}{+0.08} & \wpgcell{0.74}{+0.07} & \wpgcell{0.91}{+0.08} & \wpgcell{0.79}{+0.08} \\
\midrule
\textbf{Twitter} & mDeBERTa-v3-base (Base) & 0.66 & 0.57 & 0.69 & 0.61 \\
\rowcolor{tableanalysisrow}
 & mDeBERTa-v3-base (\textbf{Ours}) & \wpgcell{0.79}{+0.13} & \wpgcell{0.71}{+0.13} & \wpgcell{0.82}{+0.13} & \wpgcell{0.74}{+0.13} \\
 & Qwen3.5-2B (Base) & 0.77 & 0.69 & 0.85 & 0.77 \\
\rowcolor{tableanalysisrow}
 & Qwen3.5-2B (\textbf{Ours}) & \wpgcell{0.84}{+0.07} & \wpgcell{0.76}{+0.08} & \wpgcell{0.93}{+0.08} & \wpgcell{0.85}{+0.08} \\
 & Ministral-3B (Base) & 0.75 & 0.67 & 0.81 & 0.74 \\
\rowcolor{tableanalysisrow}
 & Ministral-3B (\textbf{Ours}) & \wpgcell{0.82}{+0.08} & \wpgcell{0.75}{+0.07} & \wpgcell{0.89}{+0.08} & \wpgcell{0.82}{+0.08} \\
\midrule
\textbf{eRisk25} & mDeBERTa-v3-base (Base) & 0.64 & 0.55 & 0.69 & 0.50 \\
\rowcolor{tableanalysisrow}
 & mDeBERTa-v3-base (\textbf{Ours}) & \wpgcell{0.77}{+0.13} & \wpgcell{0.69}{+0.13} & \wpgcell{0.82}{+0.13} & \wpgcell{0.63}{+0.13} \\
 & Qwen3.5-2B (Base) & 0.74 & 0.66 & 0.85 & 0.66 \\
\rowcolor{tableanalysisrow}
 & Qwen3.5-2B (\textbf{Ours}) & \wpgcell{0.82}{+0.08} & \wpgcell{0.74}{+0.08} & \wpgcell{0.93}{+0.08} & \wpgcell{0.73}{+0.08} \\
 & Ministral-3B (Base) & 0.72 & 0.66 & 0.82 & 0.62 \\
\rowcolor{tableanalysisrow}
 & Ministral-3B (\textbf{Ours}) & \wpgcell{0.81}{+0.08} & \wpgcell{0.72}{+0.07} & \wpgcell{0.90}{+0.08} & \wpgcell{0.70}{+0.08} \\
\bottomrule
\end{tabular}
\end{adjustbox}
\endgroup
\captionof{table}{In-domain matched-backbone comparison. Each Base/Ours pair
uses the same backbone; values are rounded to two decimals, and parentheses in
Ours rows report absolute gains over the matched Base row.}
\label{tab:backbone-replacement}
\end{threeparttable}
\end{minipage}\hfill
\begin{minipage}[t]{\columnwidth}
\centering
\scriptsize
\setlength{\tabcolsep}{1.7pt}
\renewcommand{\arraystretch}{0.90}
\begin{tabular*}{\linewidth}{@{\extracolsep{\fill}}llcccc@{}}
\toprule
Data & Method & Rec. & F1 & AUROC & AUPRC \\
\midrule
\tablerowshade SWDD & \textbf{Ours} & \textbf{0.8050} & \textbf{0.7490} & \textbf{0.9490} & \textbf{0.8300} \\
 & w/o Path A & 0.7251 & 0.6723 & 0.9163 & 0.7584 \\
 & w/o MoE & 0.6532 & 0.6018 & 0.8802 & 0.6934 \\
 & w/o Weak Priors & 0.7824 & 0.7287 & 0.9359 & 0.8003 \\
 & w/o Route Loss & 0.7950 & 0.7412 & 0.9427 & 0.8181 \\
 & w/o DP Dropout & 0.6889 & 0.6394 & 0.8967 & 0.7226 \\
\midrule
\tablerowshade Twitter & \textbf{Ours} & \textbf{0.8440} & \textbf{0.7630} & \textbf{0.9290} & \textbf{0.8520} \\
 & w/o Path A & 0.7775 & 0.6933 & 0.8994 & 0.7762 \\
 & w/o MoE & 0.6800 & 0.5897 & 0.8423 & 0.6689 \\
 & w/o Weak Priors & 0.8170 & 0.7339 & 0.9148 & 0.8180 \\
 & w/o Route Loss & 0.8308 & 0.7486 & 0.9215 & 0.8352 \\
 & w/o DP Dropout & 0.7350 & 0.6475 & 0.8688 & 0.7204 \\
\midrule
\tablerowshade eRisk25 & \textbf{Ours} & \textbf{0.8210} & \textbf{0.7440} & \textbf{0.9260} & \textbf{0.7350} \\
 & w/o Path A & 0.7525 & 0.6694 & 0.8911 & 0.6660 \\
 & w/o MoE & 0.6598 & 0.5702 & 0.8368 & 0.5495 \\
 & w/o Weak Priors & 0.7939 & 0.7128 & 0.9087 & 0.7002 \\
 & w/o Route Loss & 0.8067 & 0.7275 & 0.9164 & 0.7176 \\
 & w/o DP Dropout & 0.7119 & 0.6253 & 0.8652 & 0.6109 \\
\bottomrule
\end{tabular*}
\captionof{table}{In-domain ablation under the protocol of Table
\ref{tab:cross-dataset-target-tuned}. Each block evaluates on the same dataset
used for training; the complete matrix is in Table
\ref{tab:appendix-cross-dataset-ablation-full}.}
\label{tab:cross-dataset-ablation}
\end{minipage}
\vspace{0.5\baselineskip}
\end{figure*}

Table \ref{tab:cross-dataset-ablation} ablates links among
privileged evidence, deployable screening, and routing on in-domain cells; the
complete train$\rightarrow$test ablation matrix appears in Appendix
\ref{sec:appendix-full-ablation-results}. Removing dense MoE gives the largest
loss, confirming one shared state is too coarse. Path A and DP dropout are
next most important, while weak priors and route loss yield smaller but
stable drops, indicating that they shape expert allocation rather than act as
standalone predictors.

\subsection{Case Study}
\label{sec:case-study}

To examine whether WPG-MoE better handles difficult evidence layouts than previous
screening-based detectors, Figure \ref{fig:case-study} compares two anonymized
SWDD cases. \textbf{For privacy, posts are paraphrased to preserve core meaning
while removing identifiers, user details, timestamps, original sentence
structure, and exact identifiable wording.}
The sparse-evidence user shows ordinary posts with few isolated, intense
depressive cues; user-level DeCapsNet weakens them, yielding a missed/weak
output. WPG-MoE detects them by keeping sparse, mixed, and global views active.
For the episode-supported user, sleep disturbance, low mood, and withdrawal
recur without a decisive post. DeCapsNet
fragments this pattern; WPG-MoE emphasizes episode-supported/global experts for
correct output. These cases show WPG-MoE preserves evidence-layout
heterogeneity single-path screeners tend to blur.

\section{Related Work}
\label{sec:related_work}

User-level depression detection from social media has moved from holistic
profile modeling to evidence-oriented user representation
\citep{de2013predicting,coppersmith2015clpsych,song2017learning,shen2017depression,zogan2021depressionnet}.
One line of work reduces noisy posting histories by selecting or summarizing
indicator posts, modeling temporal symptom signals, or fusing heterogeneous
modalities
\citep{zeng2018topic,gui2019cooperative,zogan2021depressionnet,wang2022online,cai2023depression,ali2026overview,yu2026early}.
Another line grounds prediction in clinically meaningful structures, including
PHQ-9 questionnaires, psychiatric-scale screening, and symptom capsules
\citep{nguyen2022improving,zhang2022psychiatric,liu2024depression,bolegave2025gold,wang2025end,xu2025cnsocialdepress}.
Recent LLM-based systems further extract DSM-style symptoms, mood courses,
clinical evidence, or questionnaire responses for interpretable screening
\citep{alhamed2024using,tian2024chimed,wang2024explainable,chen2025generating,jia2025medikal,kim2025interpretable,lan2025depression,ravenda2025llms,tian2025multimodal,zhang2025explainable,gulino2026depression,thamrin2026enhancing}.
These methods improve evidence selection and interpretability, but most still
compress the selected evidence into a single user representation or pass it to a
single detector, which can blur sparse cues and episode-level patterns across
heterogeneous users.

Our work is also related to conditional computation. Mixture-of-experts (MoE)
models support specialization across examples, tasks, or modalities
\citep{jacobs1991adaptive,shazeer2017outrageously,fedus2022switch,ma2018modeling,tian2024dialogue},
and have recently been explored for mental-health prediction from textual and
non-textual social-media signals \citep{santos2023mental,dos2025mixture}.
However, existing mental-health MoE methods mainly treat experts as
representation, language, or modality specialists rather than routing them with
clinically structured evidence layouts. Learning using privileged information
(LUPI) provides a complementary view: richer signals can guide training even
when they are unavailable at deployment
\citep{vapnik2009new,lopez2015unifying,lapin2014learning}. WPG-MoE combines
these threads by using LLM-derived Path-A fields as training-only, layout-aware
weak routing priors, while inference keeps PHQ-9 screening and a shared
backbone without external LLM processing.

\section{Conclusion}
\label{sec:conclusion}

We study user-level depression detection with post-screening heterogeneity.
WPG-MoE combines dual-path evidence, weak-prior routing, and dense
experts to reduce shared-detector averaging under LUPI, with PHQ-9 screening
retained at inference. Chinese/English experiments show gains over strong
baselines with interpretable routing.
Slice and case analyses further show this gain comes from preserving sparse and
episode-supported cues under PHQ-9-only inference.

\begingroup
\def\href#1#2{#2}
\bibliography{custom}
\endgroup

\appendix

\section{Appendix}
\label{sec:appendix}

\newcommand{\appendixbox}[1]{%
\begin{center}
\begin{tcolorbox}[enhanced,breakable,colframe=RoyalBlue,colback=white,
width=0.96\linewidth,boxrule=0.6pt,arc=1mm,left=0.8mm,right=0.8mm,top=0.6mm,bottom=0.6mm]
\raggedright\scriptsize #1
\end{tcolorbox}
\end{center}}

\newcommand{\appendixwidebox}[1]{%
\begin{center}
\begin{tcolorbox}[enhanced,colframe=RoyalBlue,colback=white,
width=0.96\textwidth,boxrule=0.6pt,arc=1mm,left=0.8mm,right=0.8mm,top=0.6mm,bottom=0.6mm]
\raggedright\scriptsize #1
\end{tcolorbox}
\end{center}}

\newcommand{\appendixpromptbox}[1]{%
\begin{center}
\begin{tcolorbox}[enhanced,colframe=RoyalBlue,colback=white,
width=0.96\linewidth,boxrule=0.6pt,arc=1mm,left=0.8mm,right=0.8mm,top=0.6mm,bottom=0.6mm]
\raggedright\scriptsize #1
\end{tcolorbox}
\end{center}}

\subsection{Weak-Prior Scoring Audit}
\label{sec:appendix-weak-prior-audit}

We conduct a blinded human audit to assess whether the Qwen-derived structured
fields used by Path A provide reliable weak-prior signals. Three trained
computer-science student annotators participated after guideline training and a
qualification test. Annotators A and B independently labeled anonymized samples,
while Annotator C adjudicated only their disagreements. The annotators had
access to the anonymized text and annotation guideline, but not to Qwen scores,
model predictions, or experimental outcomes. Table
\ref{tab:appendix-weak-prior-sample} summarizes the stratified audit sample.

\paragraph{Annotator recruitment and project participation.}
The three annotators were recruited from computer-science graduate students with
prior coursework or research experience in NLP annotation. Before annotation,
they completed guideline training and a qualification test. Annotation was
conducted as part of a supervised research project. The task did not require
annotators to make clinical judgments; they labeled only predefined textual
evidence categories from anonymized packets. Annotators were allowed to pause or
stop annotation if they felt uncomfortable with mental-health-related content.

The audit sample is designed to cover both frequent and difficult evidence
patterns rather than only high-confidence self-disclosure posts. Because the
three datasets differ substantially in user count and history length, we keep all
102 eRisk25 depressed users and sample larger but still tractable subsets from
Twitter and SWDD. Twitter and SWDD users are stratified by posting volume,
maximum Path-A composite score, and preliminary evidence layout. Within each
selected user, posts are chosen to include top-ranked evidence, self-disclosure
or clinical-anchor posts, symptom/crisis/duration cues, low-score controls, and
time-random context posts; eRisk25 additionally includes early- and late-history
random posts because its user histories are much longer.
The three tables below separate sample coverage, binary post-level reliability,
and non-binary/user-level reliability, so the audit design and the scorer quality
are inspected independently.

\begin{table}[!ht]
\centering
\scriptsize
\setlength{\tabcolsep}{0pt}
\renewcommand{\arraystretch}{0.92}
\begin{tabular*}{\linewidth}{@{\extracolsep{\fill}}lrrr@{}}
\toprule
\textbf{Dataset} & \textbf{Users} & \textbf{Posts} & \textbf{Posts/User} \\
\midrule
eRisk25 & 102 & 1,020 & 10 \\
Twitter & 200 & 1,600 & 8 \\
SWDD & 400 & 3,200 & 8 \\
\midrule
Total & 702 & 5,820 & -- \\
\bottomrule
\end{tabular*}
\caption{Audit sample used for weak-prior scoring validation.}
\label{tab:appendix-weak-prior-sample}
\end{table}

\begin{table}[!ht]
\centering
\scriptsize
\setlength{\tabcolsep}{0pt}
\renewcommand{\arraystretch}{0.92}
\begin{tabular*}{\linewidth}{@{}l@{\extracolsep{\fill}}ccccc@{}}
\toprule
\textbf{Field} & \textbf{Agr.} & \textbf{$\kappa$} &
\textbf{P} & \textbf{R} & \textbf{F1} \\
\midrule
First-person & 0.885 & 0.671 & 0.780 & 0.880 & 0.827 \\
Literal self-evidence & 0.893 & 0.689 & 0.765 & 0.875 & 0.816 \\
Any symptom & 0.901 & 0.693 & 0.770 & 0.890 & 0.826 \\
Clinical anchor & 0.913 & 0.709 & 0.750 & 0.860 & 0.801 \\
Duration/frequency & 0.888 & 0.676 & 0.780 & 0.880 & 0.827 \\
Crisis any & -- & -- & 0.780 & 0.870 & 0.822 \\
Valid evidence & 0.890 & 0.681 & 0.760 & 0.900 & 0.824 \\
\bottomrule
\end{tabular*}
\caption{Post-level audit of binary weak-prior fields. Agr. and $\kappa$
measure A/B human agreement; Qwen P/R/F1 are computed against adjudicated human
labels. \textit{Crisis any} is derived from \texttt{crisis\_level} $>0$.}
\label{tab:appendix-weak-prior-post-audit}
\end{table}

\begin{table*}[!t]
\centering
\scriptsize
\setlength{\tabcolsep}{5.5pt}
\renewcommand{\arraystretch}{0.96}
\begin{tabularx}{\textwidth}{@{}>{\raggedright\arraybackslash}X>{\raggedright\arraybackslash}Xcc@{}}
\toprule
\textbf{Target} & \textbf{Metric} & \textbf{A/B} & \textbf{Qwen vs. Human} \\
\midrule
Symptom dimensions & micro/macro-F1 & 0.781 / 0.675 & 0.806 / 0.691 \\
Symptom strength & weighted $\kappa$ & 0.692 & 0.701 \\
Crisis level & weighted $\kappa$ & 0.672 & 0.685 \\
Functional impairment & weighted $\kappa$ & 0.684 & 0.664 \\
User evidence layout & exact/$\kappa$/macro-F1 & 0.754 / 0.683 / 0.692 & 0.713 / 0.654 / 0.665 \\
\bottomrule
\end{tabularx}
\caption{Audit of non-binary and user-level weak-prior fields.
\texttt{symptom\_dimensions} is a PHQ-9-aligned multi-label target, so we
report micro/macro-F1. Ordinal 0--3 fields use weighted $\kappa$. User evidence
layout is a user-level aggregation target, and Qwen vs. Human compares
Qwen-derived fields with adjudicated human labels.}
\label{tab:appendix-weak-prior-nonbinary}
\end{table*}

For the PHQ-9-aligned \texttt{symptom\_vector}, we evaluate three derived
targets: whether any symptom evidence is present (P/R/F1), which PHQ-9
dimensions are expressed (micro/macro-F1), and the overall ordinal symptom
strength (weighted $\kappa$). Cohen's $\kappa$ is chance-corrected agreement;
weighted $\kappa$ additionally penalizes larger 0--3 disagreements. Across the
audited fields, A/B agreement is substantial but not perfect, which is expected
for short, informal, and sometimes ambiguous social-media text. Qwen-to-human
scores remain consistently high for the binary fields and moderate-to-strong for
multi-label or ordinal targets, supporting their use as reliable training-time
weak-prior signals rather than as clinical labels or deploy-time inputs.

\subsection{PHQ-9 Symptom Template Inventory}
\label{sec:appendix-phq9-templates}

Path B groups template queries by the nine PHQ-9 symptom dimensions:
anhedonia, depressed mood, sleep disturbance, fatigue, appetite change, guilt
or worthlessness, concentration difficulty, psychomotor change, and self-harm
or suicidal ideation. Each template is a short natural-language query used for
semantic-similarity screening, not a diagnostic questionnaire score.
This inventory links the deployable screening branch to the same symptom
vocabulary used by Path A, while avoiding any training-only structured scorer at
inference time.

\subsection{User-Type Reference Audit}
\label{sec:appendix-user-type-audit}

We next audit the user-level evidence-layout labels induced from the weak-prior
pipeline. This audit differs from the post-level scoring audit above: annotators
inspect a complete sampled user packet and assign the dominant evidence layout
among \textit{self-disclosure}, \textit{episode-supported},
\textit{sparse-evidence}, and \textit{mixed/other}. Annotators A and B label
each packet independently, and Annotator C adjudicates only A/B disagreements.
The resulting labels are used as adjudicated reference categories for checking
the weak-prior assignment rule; they are not clinical diagnoses.

\begin{table}[H]
\centering
\scriptsize
\setlength{\tabcolsep}{1pt}
\renewcommand{\arraystretch}{0.92}
\begin{tabular*}{\linewidth}{@{\extracolsep{\fill}}llrrrrl@{}}
\toprule
\textbf{Data} & \textbf{Slice} & \textbf{Avail.} & \textbf{Audit} &
\textbf{Bndry.} & \textbf{Posts/U} & \textbf{Rule} \\
\midrule
SWDD & self-disclosure & 1,815 & 40 & 10 & 173.3 & stratified \\
SWDD & episode-supported & 1,223 & 40 & 10 & 173.3 & stratified \\
SWDD & mixed/other & 650 & 35 & 10 & 173.3 & stratified \\
SWDD & sparse-evidence & 21 & 21 & -- & 173.3 & all \\
Twitter & self-disclosure & 7 & 7 & -- & 161.1 & all \\
Twitter & episode-supported & 484 & 40 & 10 & 161.1 & stratified \\
Twitter & sparse-evidence & 271 & 35 & 10 & 161.1 & stratified \\
Twitter & mixed/other & 212 & 35 & 10 & 161.1 & stratified \\
eRisk25 & episode-supported & 71 & 71 & -- & 690.1 & all \\
eRisk25 & sparse-evidence & 5 & 5 & -- & 690.1 & all \\
eRisk25 & mixed/other & 5 & 5 & -- & 690.1 & all \\
\bottomrule
\end{tabular*}
\caption{Stratified user-level audit sample for validating weak-prior-derived
evidence-layout labels. Boundary cases denote low-margin or otherwise ambiguous
users intentionally included in the sample; -- indicates that the whole slice is
audited.}
\label{tab:appendix-user-type-sample}
\end{table}

Table~\ref{tab:appendix-user-type-sample} shows that the audit covers both
high-prevalence slices and rare slices. For rare categories, such as
\textit{sparse-evidence} users in SWDD and eRisk25, we audit all available users;
for larger categories, stratified sampling retains boundary cases to avoid
overstating reliability on easy high-margin examples. This design makes the
reference set suitable for assessing whether weak-prior grouping remains stable
under realistic ambiguity.

\begin{table}[H]
\centering
\scriptsize
\setlength{\tabcolsep}{1.8pt}
\renewcommand{\arraystretch}{0.95}
\begin{tabular*}{\linewidth}{@{\extracolsep{\fill}}lrrrrr@{}}
\toprule
\textbf{Scope} & \textbf{$N$ users} & \textbf{Raw agr.} & \textbf{Cohen's $\kappa$} &
\textbf{Kripp. $\alpha$} & \textbf{Adj. by C} \\
\midrule
Overall & 334 & 0.802 & 0.705 & 0.754 & 66 \\
SWDD & 136 & 0.801 & 0.703 & 0.752 & 27 \\
Twitter & 117 & 0.803 & 0.708 & 0.757 & 23 \\
eRisk25 & 81 & 0.803 & 0.701 & 0.751 & 16 \\
\bottomrule
\end{tabular*}
\caption{Agreement statistics for the user-level reference audit. Raw
agreement, Cohen's $\kappa$, and Krippendorff's $\alpha$ are computed from the
independent A/B labels; C adjudicates only A/B disagreements.}
\label{tab:appendix-user-type-agreement}
\end{table}

The agreement results in Table~\ref{tab:appendix-user-type-agreement} are
consistent across datasets despite differences in language, platform, and
history length. Overall A/B agreement reaches 0.802, with Cohen's
$\kappa=0.705$ and Krippendorff's $\alpha=0.754$, indicating substantial
chance-corrected reliability for a four-way evidence-layout judgment. The 66
adjudicated disagreements are concentrated in boundary cases, where direct
self-disclosure, repeated episode evidence, and mixed evidence can overlap; this
is precisely why WPG-MoE uses these categories as soft routing tendencies rather
than fixed clinical subtypes.

\begin{table*}[!t]
\centering
\small
\setlength{\tabcolsep}{3pt}
\renewcommand{\arraystretch}{1.04}
\begin{tabular*}{\textwidth}{@{\extracolsep{\fill}}lrrrrr@{}}
\toprule
\textbf{Ref. slice} & \textbf{Sup.} & \textbf{P} &
\textbf{R} & \textbf{F1} & \textbf{Confusion} \\
\midrule
self-disclosure & 47 & 0.609 & 0.596 & 0.602 & episode-supported \\
episode-supported & 151 & 0.861 & 0.861 & 0.861 & mixed/other \\
sparse-evidence & 61 & 0.613 & 0.623 & 0.618 & mixed/other \\
mixed/other & 75 & 0.827 & 0.827 & 0.827 & episode-supported \\
\midrule
Overall & 334 & \multicolumn{4}{c@{}}{Acc. = 0.772;\hspace{1.5em} macro-F1 = 0.727;\hspace{1.5em} weighted-F1 = 0.772} \\
\bottomrule
\end{tabular*}
\caption{Weak-prior user assignment against adjudicated reference labels.
Per-slice precision, recall, and F1 evaluate the deterministic assignment rule
used to construct user evidence-layout labels; main confusion reports the most
frequent non-matching predicted slice.}
\label{tab:appendix-user-type-assignment}
\end{table*}

Table~\ref{tab:appendix-user-type-assignment} indicates that the deterministic
weak-prior assignment rule recovers the adjudicated reference labels with 0.772
accuracy and 0.727 macro-F1. The strongest alignment appears for
\textit{episode-supported} and \textit{mixed/other} users, while
\textit{self-disclosure} and \textit{sparse-evidence} are more often confused
with neighboring layouts. These errors are interpretable: self-disclosure posts
can also occur inside repeated episodes, and sparse high-risk users often border
on mixed evidence when a few additional moderate posts are present. The audit
therefore supports the quality of the weak-prior grouping while reinforcing the
paper's design choice to use it as soft privileged supervision rather than as a
hard target at inference time.

\subsection{User-Type Examples}

Table \ref{tab:appendix-user-types} shows one representative eRisk user for
each coarse evidence tendency. To avoid surfacing raw user handles, we replace
the original identifiers with E1--E3 while keeping the underlying post excerpts
and derived weak-prior patterns unchanged.
The purpose of these examples is to make the routing targets in Section
\ref{sec:approach} inspectable: the same depressed label can be supported by a
direct self-report, a temporally repeated episode, or a small number of intense
posts. The rows therefore illustrate evidence layouts rather than clinical
subtypes, and they explain why WPG-MoE uses soft tendency-specific views together
with a global fallback expert.

\begin{table*}[!t]
\centering
\footnotesize
\setlength{\tabcolsep}{5pt}
\renewcommand{\arraystretch}{1.08}
\begin{tabularx}{\textwidth}{@{}>{\raggedright\arraybackslash}p{0.18\textwidth}>{\raggedright\arraybackslash}X>{\raggedright\arraybackslash}X@{}}
\toprule
\textbf{Type} & \textbf{Representative excerpts} & \textbf{Why it matches the channel} \\
\midrule
Self-disclosure (E1) &
``I am finally taking a firm decision to get help after 10 years of major
depression \ldots'' &
This user contains direct first-person disclosure of diagnosis and treatment
seeking. The self-disclosure prior is therefore high, while the user has only a
short supporting block, so the key signal is explicit self-report rather than
long-range aggregation. \\

Episode-supported evidence (E2) &
``My PHQ-9 is down to a 20 \ldots it was 25+ for the last two years''; ``2 Day
migraine \ldots Depression x10 \ldots I just want to die.'' &
This user has three evidence blocks spanning 29, 48, and 24 days. The decision
is supported by repeated symptoms across temporally linked posts rather than by
a single disclosure post, which is exactly the pattern targeted by the
episode-supported channel. \\

Sparse high-risk evidence (E3) &
``I just feel like vanishing right now''; ``My mother doesn't believe that I'm
depressed despite having been diagnosed with it \ldots'' &
This user has only one short evidence block and relatively few risk posts, but
several of them are intense enough to raise the sparse-evidence prior. The key
signal is concentrated in a small number of high-impact posts rather than in a
dense episode-like cluster. \\
\bottomrule
\end{tabularx}
\caption{Representative user types from the processed eRisk data. The examples
are anonymized evidence-layout cases rather than diagnostic subtypes; the
rationale column states which weak-prior channel each pattern supports.}
\label{tab:appendix-user-types}
\end{table*}

\subsection{Implementation Details and Hyperparameters}
\label{sec:appendix-implementation-details}

This subsection records the code-level choices that instantiate the operators
in Section~\ref{sec:approach}. We include only settings that affect
reproducibility; the external Path-A annotation schema is shown separately in
Appendix~\ref{sec:appendix-llm-prompt}.

\paragraph{Model size and compute budget.}
The default WPG-MoE configuration uses Qwen3.5-2B as the deployable backbone
(approximately 2B parameters). Across backbone-replacement and baseline
experiments, we also evaluate BERT-base (approximately 110M parameters),
mDeBERTa-v3-base (approximately 278M parameters), Qwen3.5-0.8B,
Ministral-3B, and Ministral-8B. The measured training time below refers to the
default Qwen3.5-2B WPG-MoE runs. Training uses data parallelism on four NVIDIA
A100 GPUs. One source-dataset run takes about 4 hours for SWDD and about
2 hours for Twitter or eRisk25.

\begin{table}[!t]
\centering
\scriptsize
\setlength{\tabcolsep}{2pt}
\renewcommand{\arraystretch}{0.96}
\begin{tabularx}{\linewidth}{@{}>{\raggedright\arraybackslash}p{0.26\linewidth}>{\raggedright\arraybackslash}X@{}}
\toprule
\textbf{Component} & \textbf{Implementation setting} \\
\midrule
Template encoder & \texttt{gte-small-zh} for SWDD; \texttt{all-MiniLM-L6-v2}
for Twitter/eRisk; normalized embeddings. \\
\addlinespace[3pt]
Backbone & Qwen3.5-2B in the main runs; automatic pooling uses the last token. \\
\addlinespace[3pt]
Candidate/history caps & 32 risk candidates; eight chronological history
segments; 60\% history coverage; 12 posts per segment in full-parameter Qwen
configs. \\
\addlinespace[3pt]
Training perturbations & Risk-source swap 0.5, metadata drop 0.5, block drop
0.4, prior drop 0.3, post drop 0.3; Twitter/eRisk configs use template-source
swap 1.0. \\
\addlinespace[3pt]
Optimizer & AdamW; head learning rate $10^{-4}$; encoder learning rate
$10^{-5}$ in full-parameter Qwen configs; weight decay 0.01; gradient clipping
1.0. \\
\addlinespace[3pt]
Training schedule & Stage-D expert warm start for 3 epochs; Stage-E joint
training for up to 3 epochs with validation-F1 early stopping; effective
Stage-E batch size 8 in full-parameter Qwen configs. \\
\addlinespace[3pt]
Loss weights & $\alpha=0.3$, $\beta=0.2$, $\gamma=0.15$; entropy weight decays
from 0.1 to 0.02 with a cosine schedule. \\
\addlinespace[3pt]
Routing loss & Applied only when the largest weak prior is at least 0.6 and
exceeds the second largest by at least 0.1; KL aligns the first three gate
weights with normalized weak priors. \\
\addlinespace[3pt]
Balance/entropy losses & Balance uses batch-level importance and sharpened
load with temperature 0.1; entropy is added as negative gate entropy to
encourage early routing diversity. \\
\bottomrule
\end{tabularx}
\caption{Reproducibility-critical implementation settings confirmed from the
training and inference code.}
\label{tab:appendix-implementation-hyperparameters}
\end{table}

\paragraph{Candidate scores and budgets.} For a user with $n$ posts, both
Path A and Path B keep the top $K(n)$ posts, where
\[
K(n)=
\begin{cases}
\lceil 0.125n \rceil, & n\geq 160,\\
20, & 20\leq n<160,\\
n, & n<20.
\end{cases}
\]
Path A maps structured LLM fields to a weak evidence ranker, not a tuned
post-level classifier. Let $e_i^A=[q_i,c_i,a_i,d_i,r_i,u_i]\in[0,1]^6$ collect
normalized symptom, crisis, clinical-anchor, duration, confidence, and literal
self-disclosure cues, where $q_i$ summarizes PHQ-9 symptom intensity and
coverage from $v_i\in\{0,1,2,3\}^{9}$. We compute
\[
\begin{aligned}
s_i^A&=\mathrm{clip}_{[0,1]}\!\left((w^A)^\top e_i^A\right),\\
w^A&\geq 0,\qquad \|w^A\|_1=1 .
\end{aligned}
\]
where $w^A$ is a fixed symptom-heavy vector, not selected on validation or test
targets. The score only constructs training-time weak priors and is removed at
inference.
Path B embeds each post and the PHQ-9 template inventory with normalized
sentence embeddings. For dimension $d$, the dimension score is the maximum
cosine similarity to that dimension's three templates. The template risk score
uses a fixed pooling rule over the strongest PHQ-9 dimensions,
$s_i^B=\mathrm{Pool}_{\mathrm{PHQ}}(\{b_{i,d}\}_{d=1}^{D})$; matched dimensions
are retained as lightweight metadata. The final model input uses at most 32
candidate posts in the Qwen full-parameter configurations.

\paragraph{Evidence blocks and weak priors.} Evidence blocks are built
only for depressed training users with Path-A scores. A post is eligible when
it is first-person, literal self-evidence, and passes a fixed weak-evidence
threshold. Eligible posts are ordered by timestamp and adjacent posts within a
short temporal window are merged. Blocks are scored by monotone pooling over
normalized block evidence:
\[
B=\mathrm{Pool}_{\mathrm{blk}}
\bigl(\tilde n_b,\tilde \ell_b,\tilde m_b,d_b,\tilde f_b,\bar r_b\bigr),
\]
where the tilded variables normalize block post count, span, symptom coverage,
and impairment; $d_b$ indicates duration and $\bar r_b$ is average confidence.
Each block stores representative posts, and each user keeps only top-ranked
blocks.

The implemented weak prior vector is
$\pi=[\pi_{\mathrm{sd}},\pi_{\mathrm{ep}},\pi_{\mathrm{sp}}]$. The
self-disclosure prior averages per-post evidence from current self-claims,
clinical anchors, current literal self-evidence, and confidence. The
episode-supported prior is computed from the best block using its post count,
span, symptom coverage, duration support, and functional impairment. The sparse
evidence prior is activated when neither self-disclosure nor episode evidence
dominates and the user has only a few high-scoring posts; it combines the top
composite scores and average confidence. The crisis score is the maximum
Path-A crisis level and is normalized before being passed to the model.
At inference time, Path A, episode blocks, weak priors, and crisis annotations
are removed; Path B candidates and lightweight statistics remain.

\paragraph{Representation and routing.} The shared
encoder produces post representations that are aggregated into five views:
$z_{\mathrm{sd}}$ attends over all risk candidates, $z_{\mathrm{ep}}$ attends
over block posts and falls back to risk candidates when blocks are absent,
$z_{\mathrm{sp}}$ attends over the first three risk candidates,
$z_{\mathrm{mix}}$ mean-pools risk candidates, and $z_g$ applies temporal
self-attention to eight global-history segments plus a statistics projection.
The attention-pooling modules are separate linear scoring heads, so the three
evidence-oriented views learn different pooling weights. The gate is a
two-layer MLP with hidden size 256 and dropout 0.1 over
$[z_{\mathrm{sd}};z_{\mathrm{ep}};z_{\mathrm{sp}};z_{\mathrm{mix}};z_g;\pi;c;\mathrm{stats}]$.
All five experts are evaluated for every user. Each expert receives one view
concatenated with a 10-dimensional projected metadata vector and uses
Linear$(d+10,512)$, GELU, dropout 0.1, and Linear$(512,256)$, where $d$ is the
encoder hidden size; the MoE state is
the dense weighted sum of expert outputs. The auxiliary scoring head maps
$[x_i;h;g]$ through an MLP and sigmoid for training-time evidence supervision.

\subsection{Training-Time External LLM Scoring Prompt}
\label{sec:appendix-llm-prompt}

Path A is built by an offline single-post scoring script that feeds
\texttt{tweet\_index}, \texttt{posting\_time}, and \texttt{text} into a
schema-constrained LLM annotation interface. The repository prompt is written in
Chinese; for readability, we show a faithful English translation below. The
structured output is then used for composite post scoring, evidence-block
construction, and weak-prior induction. This external prompt defines the
training-time privileged scorer only; it is distinct from the deployable
Qwen3.5-2B backbone used in the main model.

\appendixbox{
\textbf{System prompt (English translation).} You are a data annotation
assistant for mental-health research. Your task is \emph{post-level evidence
extraction}, not clinical diagnosis, not PHQ-9 total-score calculation, and not
counseling. Judge only from the given post text and rely strictly on explicit
textual evidence rather than speculation.

\medskip
\textbf{Return the following fields.}
\begin{itemize}
\item \texttt{first\_person} (boolean): whether the post is written from the
author's own perspective.
\item \texttt{literal\_self\_evidence} (boolean): whether the post is a direct
self-report rather than quotation, news, jokes, lyrics, metaphor, or
third-person discussion.
\item \texttt{symptom\_vector}: a 9-dimensional PHQ-9-aligned symptom evidence
vector with values in \{0,1,2,3\}.
\item \texttt{crisis\_level} (0/1/2/3): suicide, self-harm, or acute collapse
risk.
\item \texttt{duration}: whether the post contains a duration or frequency cue;
if a day span is explicit, return it as \texttt{hint\_span\_days}, otherwise
\texttt{null}.
\item \texttt{functional\_impairment} (0/1/2/3): whether the post indicates
functional impairment.
\item \texttt{clinical\_context}: includes
\texttt{disease\_mention\_type}=\{none, generic\_topic, self\_history,
current\_self\_claim\} and a subset of
\texttt{anchor\_types}=\{diagnosis, doctor\_visit, psychiatry,
hospitalization, medication, follow\_up, therapy\}.
\item \texttt{temporality}=\{current, past, recovery, unclear\}.
\item \texttt{confidence} (0.0--1.0): confidence that the post contains genuine
mental-health evidence rather than quotation, humor, or hearsay.
\end{itemize}

\textbf{Constraints.} This is single-post evidence extraction rather than
user-level diagnosis. The values in \texttt{symptom\_vector} denote \emph{textual
evidence strength} rather than PHQ-9 two-week frequency. Absence of evidence
means lack of information, not counter-evidence. The \texttt{posting\_time}
string must be copied verbatim into the output. Return valid JSON only, with no
extra text.
}

\appendixbox{
\textbf{Structured output template.}

\medskip
{\ttfamily
\{\\
\quad "tweet\_index": <int>,\\
\quad "posting\_time": "<original string>",\\
\quad "first\_person": <bool>,\\
\quad "literal\_self\_evidence": <bool>,\\
\quad "symptom\_vector": \{\\
\qquad "depressed\_mood": 0/1/2/3,\\
\qquad "anhedonia": 0/1/2/3,\\
\qquad "sleep": 0/1/2/3,\\
\qquad "fatigue": 0/1/2/3,\\
\qquad "appetite\_or\_weight": 0/1/2/3,\\
\qquad "worthlessness\_or\_guilt": 0/1/2/3,\\
\qquad "concentration": 0/1/2/3,\\
\qquad "psychomotor": 0/1/2/3,\\
\qquad "suicidal\_ideation": 0/1/2/3\\
\quad \},\\
\quad "crisis\_level": 0/1/2/3,\\
\quad "duration": \{"has\_hint": <bool>, "hint\_span\_days": <int|null>\},\\
\quad "functional\_impairment": 0/1/2/3,\\
\quad "clinical\_context": \{\\
\qquad "disease\_mention\_type":\\
\qquad\qquad "<one of: none, generic\_topic,",\\
\qquad\qquad "self\_history, current\_self\_claim>",\\
\qquad "anchor\_types": ["diagnosis", "..."]\\
\quad \},\\
\quad "temporality": "current|past|recovery|unclear",\\
\quad "confidence": <float>\\
\}
}
}

\appendixbox{
\textbf{Illustrative scored example.}

\medskip
{\ttfamily
\{\\
\quad "tweet\_index": 17,\\
\quad "posting\_time": "2024-03-12 21:14:05",\\
\quad "first\_person": true,\\
\quad "literal\_self\_evidence": true,\\
\quad "symptom\_vector": \{\\
\qquad "depressed\_mood": 3, "anhedonia": 2, "sleep": 2,\\
\qquad "fatigue": 2, "appetite\_or\_weight": 0,\\
\qquad "worthlessness\_or\_guilt": 2, "concentration": 1,\\
\qquad "psychomotor": 0, "suicidal\_ideation": 1\\
\quad \},\\
\quad "crisis\_level": 1,\\
\quad "duration": \{"has\_hint": true, "hint\_span\_days": 14\},\\
\quad "functional\_impairment": 2,\\
\quad "clinical\_context": \{\\
\qquad "disease\_mention\_type": "current\_self\_claim",\\
\qquad "anchor\_types": ["doctor\_visit", "medication"]\\
\quad \},\\
\quad "temporality": "current",\\
\quad "confidence": 0.94\\
\}
}
}

\subsection{SWDD Label-Noise Cases}
\label{sec:appendix-swdd-noise-examples}

\begin{table}[H]
\centering
\scriptsize
\setlength{\tabcolsep}{2.5pt}
\renewcommand{\arraystretch}{0.90}
\begin{tabularx}{\linewidth}{@{}l>{\raggedright\arraybackslash}Xrr@{}}
\toprule
Stage & Interpretation & $n$ & Note \\
\midrule
Raw & Self-reported (raw 1) & 1,434 & 38.6\% \\
Raw & Non-self (raw 0) & 2,277 & 61.4\% \\
Audit & Raw 0 $\rightarrow$ corr. 1 & 399 & 17.5\% \\
Audit & Raw 1 $\rightarrow$ corr. 0 & 16 & 1.1\% \\
Corrected & Self-reported (corr. 1) & 1,817 & 49.0\% \\
Corrected & Non-self (corr. 0) & 1,894 & 51.0\% \\
\bottomrule
\end{tabularx}
\caption{SWDD self-report audit summary. Raw rows show the original
\texttt{self\_reported} flag, audit rows show the two correction directions, and
corrected rows give the final slice counts used in our SWDD analyses.}
\label{tab:swdd-noise-audit}
\end{table}

Our audit distinguishes two failure modes in the raw SWDD
\texttt{self\_reported} flag: missed self-disclosure among raw negatives, and
raw positives whose retained posts do not provide clear first-person
self-disclosure evidence. Table~\ref{tab:swdd-noise-audit} summarizes the
final corrected counts. Starting from 1,434 raw positives and 2,277 raw
negatives, the forward mismatch screen moves 399 raw negatives with direct
first-person self-disclosure evidence to corrected self-reported, while the
reverse screen moves the conservative Category-A subset of 16 raw positives,
whose retained posts contain no depression, anxiety, medication, or treatment
keywords, to corrected non-self. The final audit yields 1,817 corrected
self-reported users and 1,894 corrected non-self users. Table
\ref{tab:appendix-swdd-noise} gives representative audit patterns behind these
two correction directions.

\begin{table*}[tbp]
\centering
\scriptsize
\setlength{\tabcolsep}{3pt}
\renewcommand{\arraystretch}{0.90}
\begin{tabularx}{\textwidth}{@{}l l X X@{}}
\toprule
\textbf{Case} & \textbf{Raw flag} & \textbf{Representative evidence} & \textbf{Audit rationale} \\
\midrule
S1 & \texttt{self\_reported=False} &
``I thought I had already passed the worst stage \ldots until I was diagnosed
with depression two days ago \ldots I do not know whether staying on medication
will help, or whether stopping it will make me relapse.'' &
This post contains direct first-person diagnosis, medication, and treatment
markers. It therefore contradicts the raw negative self-report flag and is
reassigned to the self-disclosure slice under our audit protocol. \\

S2 & \texttt{self\_reported=True} &
Category A retained histories contain no depression, anxiety, medication, or
treatment keywords after de-identification. &
This conservative raw-positive pattern provides no retained textual evidence
for mental-health self-disclosure. Such users motivate reverse screening of raw
positives rather than blind acceptance of the original \texttt{self\_reported}
flag. \\
\bottomrule
\end{tabularx}
\caption{Representative SWDD label-audit patterns. S1 illustrates a raw-negative
user corrected to self-disclosure, whereas S2 summarizes the conservative
Category-A raw-positive pattern whose retained posts contain no
depression/anxiety/medication/treatment keywords.}
\label{tab:appendix-swdd-noise}
\end{table*}

\subsection{Diagnostic Split Details}
\label{sec:appendix-diagnostic-splits}

\paragraph{Human-labeled diagnostic.}
The main slice-level diagnostic uses human evidence-layout labels assigned from
raw user packets. During annotation, each packet exposes only an anonymous user
ID, original post text, relative temporal order, and necessary context posts.
Annotators do not see Path-A scores, weak-prior values, model predictions,
expert weights, original rule-derived labels, dataset names, raw account IDs, or
train/validation/test splits.

Table \ref{tab:appendix-human-diagnostic-pool} gives the full diagnostic pool.
Positive users are sampled from all three datasets to obtain enough rare
\textit{sparse-evidence} and \textit{mixed/other} cases; controls are fixed
across evidence-layout slices and stratified by history length.

\begin{table}[!htbp]
\centering
\scriptsize
\setlength{\tabcolsep}{2pt}
\renewcommand{\arraystretch}{1.04}
\begin{tabular*}{\linewidth}{@{\extracolsep{\fill}}llrrrr@{}}
\toprule
\textbf{Data} & \textbf{Group} & \textbf{Total} & \textbf{Short} &
\textbf{Med.} & \textbf{Long} \\
\midrule
SWDD & Dep. & 180 & 45 & 90 & 45 \\
SWDD & Ctrl. & 900 & 225 & 450 & 225 \\
Twitter & Dep. & 120 & 30 & 60 & 30 \\
Twitter & Ctrl. & 600 & 150 & 300 & 150 \\
eRisk25 & Dep. & 60 & 15 & 30 & 15 \\
eRisk25 & Ctrl. & 300 & 75 & 150 & 75 \\
\midrule
Total & Dep. & 360 & 90 & 180 & 90 \\
Total & Ctrl. & 1,800 & 450 & 900 & 450 \\
\bottomrule
\end{tabular*}
\caption{Human-labeled diagnostic pool. The positive side is annotated for
evidence layout; the control side is used as the fixed negative pool for all
slice evaluations.}
\label{tab:appendix-human-diagnostic-pool}
\end{table}

Table \ref{tab:appendix-human-layout-distribution} reports the adjudicated
evidence-layout distribution. Disputes count users for which Annotators A and B
assigned different primary labels before Annotator C adjudication.

\begin{table}[!htbp]
\centering
\scriptsize
\setlength{\tabcolsep}{1.2pt}
\renewcommand{\arraystretch}{0.94}
\resizebox{\linewidth}{!}{%
\begin{tabular}{lrrrrrrl}
\toprule
\textbf{Layout} & \textbf{A} & \textbf{B} & \textbf{Final} &
\textbf{\%} & \textbf{Disp.} & \textbf{Disp. \%} & \textbf{Main confusion} \\
\midrule
self-disclosure & 173 & 164 & 170 & 47.2 & 24 & 14.1 & episode-supported \\
episode-supported & 94 & 101 & 96 & 26.7 & 22 & 22.9 & self-disclosure \\
sparse-evidence & 22 & 26 & 23 & 6.4 & 8 & 34.8 & mixed/other \\
mixed/other & 46 & 43 & 45 & 12.5 & 14 & 31.1 & episode-supported \\
unable-to-judge & 25 & 26 & 26 & 7.2 & 4 & 15.4 & mixed/other \\
\midrule
Overall & 360 & 360 & 360 & 100.0 & 72 & 20.0 & -- \\
\bottomrule
\end{tabular}}
\caption{Adjudicated evidence-layout distribution for the positive diagnostic
pool. Final labels are used for Figure \ref{fig:exp1-slice-auprc};
unable-to-judge users are excluded from type-specific evaluation.}
\label{tab:appendix-human-layout-distribution}
\end{table}

\paragraph{Human annotation guideline excerpt.}
We write a dedicated user-level annotation guideline before the annotation. The
excerpt below preserves the decision rules used by the annotators.

\appendixbox{
\textbf{Purpose.} The annotation unit is a user, not a single post. Annotators do
not re-diagnose whether the user is depressed; they judge the dominant form in
which depression-risk evidence appears across the user's social-media history.

\medskip
\textbf{Visible and prohibited information.} Annotators only see an anonymous
user ID, the original post text in the packet, relative temporal order
(\textit{early}, \textit{middle}, \textit{late}, or \textit{single}), and
necessary context posts. They must not see or infer Path-A LLM scores,
weak-prior values, model predictions, expert or gate weights, original
rule-derived labels, structured fields generated by Path A, dataset names, raw
account IDs, or split membership.

\medskip
\textbf{General principles.} The primary label is the most stable evidence
layout that explains the user's risk evidence. A secondary label records a clear
but non-dominant second layout; otherwise it is \texttt{none}. Annotators should
not treat ordinary complaints, jokes, lyrics, reposts, film lines, or other
people's experiences as the user's own depression evidence. First-person
evidence, sustained repetition, cross-time recurrence, functional impairment,
treatment information, and crisis intensity are the main decision cues. If the
evidence is insufficient, annotators may choose \textit{unable-to-judge} instead
of forcing a category.
}

\appendixbox{
\textbf{Label definitions.}
\textit{Self-disclosure} applies when the user explicitly says in the first
person that they have depression, have been diagnosed, receive psychiatric or
psychological treatment, are hospitalized or followed up, or take antidepressant
or related psychiatric medication. Mere phrases such as ``I feel depressed
today,'' public information, lyrics, reposts, or descriptions of others are
excluded.
\textit{Episode-supported} applies when there is no clear diagnostic or
treatment self-disclosure, but depressive symptoms recur across multiple posts
and time points. Typical cues include persistent low mood, anhedonia, sleep or
appetite change, fatigue, guilt or worthlessness, impaired concentration, social
withdrawal, work or study impairment, and long-lasting hopelessness.
\textit{Sparse-evidence} applies when most of the history is unrelated or weak,
but a few posts contain high-intensity crisis evidence such as self-harm,
suicidal ideation, extreme despair, or severe collapse, without a sustained
cross-time symptom chain or dominant treatment self-disclosure.
\textit{Mixed/other} applies when several layouts are present with similar
strength, or when the boundary is unclear because self-disclosure, sustained
symptoms, and high-risk posts overlap. It also covers evidence that is related
to depression risk but does not stably match the previous three layouts.
\textit{Unable-to-judge} applies when the packet lacks reliable user-owned
evidence, contains too little text, is dominated by unreadable noise, links,
lyrics, reposts, advertisements, or third-person descriptions, or only contains
ordinary negative emotion without duration, crisis strength, treatment context,
or first-person evidence.
}

\appendixbox{
\textbf{Conflict rules.} Explicit first-person diagnosis, treatment, or
medication self-disclosure takes priority as the primary label; sustained
symptoms may become the secondary label. Without self-disclosure, repeated
multi-time symptoms lead to \textit{episode-supported}. A few isolated
high-intensity crisis posts lead to \textit{sparse-evidence}. If two or three
patterns are similarly strong, the primary label is \textit{mixed/other}. Lyrics,
jokes, sarcasm, reposts, and third-person statements should not support a
high-confidence primary label unless context clearly points to the user.

\medskip
\textbf{Output and quality control.} Each user receives
\texttt{annotation\_user\_id}, \texttt{primary\_label},
\texttt{secondary\_label}, \texttt{confidence},
\texttt{evidence\_post\_indices}, \texttt{evidence\_summary}, and
\texttt{decision\_rationale}. Annotators A and B label independently. Annotator
C adjudicates only A/B disagreements and does not alter A/B agreements. The
reported audit includes raw agreement, chance-corrected agreement, per-layout
counts, dispute rates, and the adjudicated label distribution.
}

\begin{figure*}[!t]
\centering
\includegraphics[width=\textwidth]{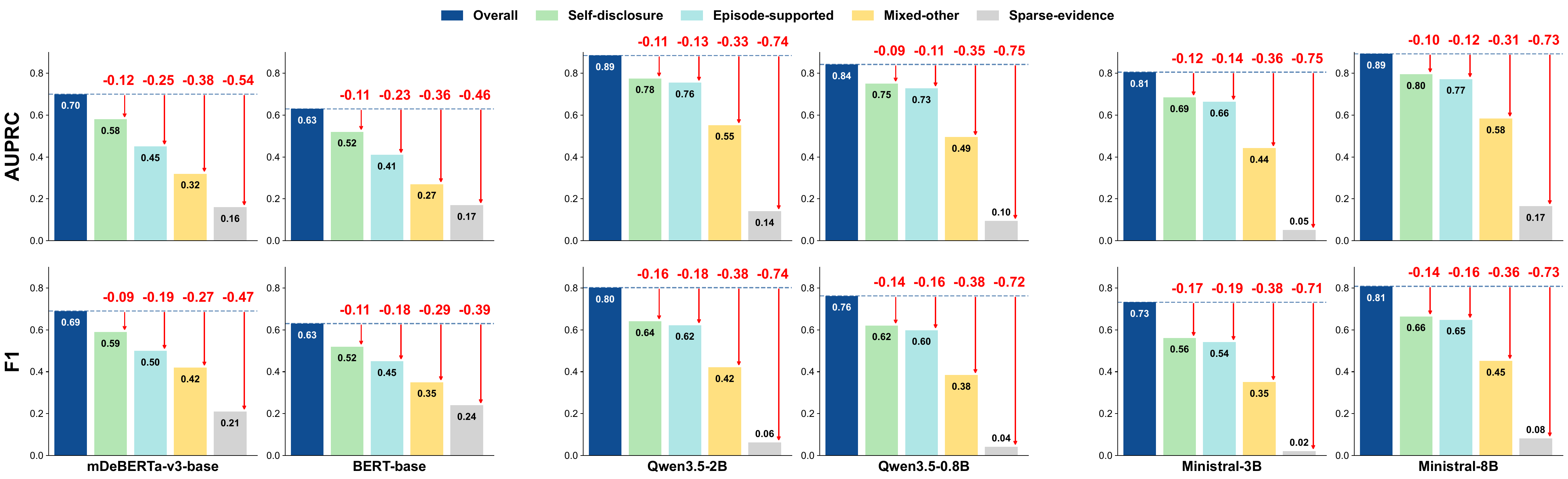}
\caption{Large-scale rule-derived slice diagnostic on corrected SWDD. AUPRC and
F1 are reported for flat detectors under the tendency and block rules, with
controls fixed across evidence-layout slices.}
\label{fig:appendix-rule-slice-auprc}
\end{figure*}

\paragraph{Large-scale rule-derived diagnostic.}
Figure \ref{fig:appendix-rule-slice-auprc} reports a larger corrected-SWDD
diagnostic. Corrected positive users are partitioned into self-disclosure,
episode-supported, sparse-evidence, and mixed/other layouts with the tendency
and block rules in Section \ref{sec:approach}, while controls are held fixed
under the same train/validation/test protocol. This analysis runs the same
difficulty check on a larger rule-derived SWDD split. The ordering is
consistent:
\textit{self-disclosure} and \textit{episode-supported} are easier,
\textit{mixed/other} is harder, and \textit{sparse-evidence} is the most
difficult.

Tables \ref{tab:appendix-slice-splits} and
\ref{tab:appendix-controlled-mixing-splits} give the corrected-SWDD split counts
and controlled-mixing construction used in Figure
\ref{fig:controlled-mixing-routing}. The sparse slice is retained for diagnostic
evaluation but not used as a primary controlled-mixing target because its
training count is too small for stable mixing curves.

The controlled-mixing table isolates whether adding non-target positive users
helps or dilutes a target evidence layout. For each target, validation and test
sets remain fixed, controls remain fixed at 7,500 training users, and only the
number of additional positive users from other layouts is varied. This
construction separates evidence-layout compatibility from simple changes in test
composition.

\begin{table*}[!t]
\centering
\footnotesize
\setlength{\tabcolsep}{0pt}
\renewcommand{\arraystretch}{1.04}
\begin{tabular*}{\textwidth}{@{\extracolsep{\fill}}lrrrrrrr@{}}
\toprule
Split & Pos. & Ctrl. & Self & Epis. & Sparse & Mixed & Total \\
\midrule
Train & 1,500 & 7,500 & 734 & 494 & 9 & 263 & 9,000 \\
Val. & 300 & 1,500 & 147 & 99 & 2 & 52 & 1,800 \\
Test & 1,000 & 5,000 & 489 & 330 & 5 & 176 & 6,000 \\
\bottomrule
\end{tabular*}
\caption{Corrected-SWDD split counts used for the slice-level
heterogeneity diagnostic. Counts are computed after label correction and the
controlled 80/10/10 holdout; Self/Epis./Sparse/Mixed partition positive users.}
\label{tab:appendix-slice-splits}
\end{table*}

\begin{table*}[!t]
\centering
\footnotesize
\setlength{\tabcolsep}{0pt}
\renewcommand{\arraystretch}{1.0}
\begin{tabular*}{\textwidth}{@{\extracolsep{\fill}}rcccccc@{}}
\toprule
\textbf{Ratio} &
\multicolumn{3}{c}{Self-disclosure target} &
\multicolumn{3}{c}{Episode target} \\
\cmidrule(lr){2-4}\cmidrule(lr){5-7}
 & Extra & Train pos. & Total & Extra & Train pos. & Total \\
\midrule
0\% & 0 & 734 & 8,234 & 0 & 494 & 7,994 \\
10\% & 77 & 811 & 8,311 & 73 & 567 & 8,067 \\
20\% & 153 & 887 & 8,387 & 147 & 641 & 8,141 \\
30\% & 230 & 964 & 8,464 & 220 & 714 & 8,214 \\
40\% & 306 & 1,040 & 8,540 & 294 & 788 & 8,288 \\
50\% & 383 & 1,117 & 8,617 & 367 & 861 & 8,361 \\
60\% & 460 & 1,194 & 8,694 & 440 & 934 & 8,434 \\
70\% & 536 & 1,270 & 8,770 & 514 & 1,008 & 8,508 \\
80\% & 613 & 1,347 & 8,847 & 587 & 1,081 & 8,581 \\
100\% & 766 & 1,500 & 9,000 & 734 & 1,228 & 8,728 \\
\bottomrule
\end{tabular*}
\caption{Controlled-mixing training splits. Extra denotes non-target positives
added to the target-slice training set; controls are fixed at 7,500 and
validation/test sets are held fixed for each target slice.}
\label{tab:appendix-controlled-mixing-splits}
\end{table*}

\subsection{Full Backbone Replacement Results}
\label{sec:appendix-full-backbone-replacement}

Table \ref{tab:appendix-backbone-replacement-full} gives the full
train$\rightarrow$test matrix corresponding to Table
\ref{tab:backbone-replacement}. The main paper keeps only the diagonal
in-domain cells to reserve space, while the complete matrix shows that the same
matched-backbone comparison also holds under cross-dataset transfer. Across
mDeBERTa-v3-base, Qwen3.5-2B, and Ministral-3B, WPG-MoE improves over the
corresponding Base classifier in every train$\rightarrow$test cell, indicating
that the gains come from the weak-prior-guided routing design rather than from a
specific pretrained backbone.

\begin{table*}[!t]
\centering
\scriptsize
\setlength{\tabcolsep}{3pt}
\renewcommand{\arraystretch}{0.82}
\resizebox{\textwidth}{!}{
\begin{tabular}{llcccc@{\hspace{8pt}}cccc@{\hspace{8pt}}cccc}
\toprule
Train on & Method & \multicolumn{4}{c}{Test on SWDD} & \multicolumn{4}{c}{Test on Twitter} & \multicolumn{4}{c}{Test on eRisk25} \\
\cmidrule(lr){3-6}\cmidrule(lr){7-10}\cmidrule(lr){11-14}
 &  & Rec. & F1 & AUROC & AUPRC & Rec. & F1 & AUROC & AUPRC & Rec. & F1 & AUROC & AUPRC \\
\midrule
SWDD & mDeBERTa-v3-base (Base) & 0.6415 & 0.5781 & 0.7304 & 0.5955 & 0.5842 & 0.4815 & 0.5921 & 0.4208 & 0.6055 & 0.5126 & 0.6384 & 0.3891 \\
\rowcolor{oursgray}
 & mDeBERTa-v3-base (\textbf{Ours}) & \textbf{0.7612} & \textbf{0.7104} & \textbf{0.8631} & \textbf{0.7392} & \textbf{0.7245} & \textbf{0.6087} & \textbf{0.7188} & \textbf{0.5524} & \textbf{0.7369} & \textbf{0.6432} & \textbf{0.7701} & \textbf{0.5175} \\
 & Qwen3.5-2B (Base) & 0.7391 & 0.6745 & 0.8705 & 0.7521 & 0.7165 & 0.5932 & 0.7495 & 0.5824 & 0.7284 & 0.6351 & 0.8021 & 0.5532 \\
\rowcolor{oursgray}
 & Qwen3.5-2B (\textbf{Ours}) & \textbf{0.8050} & \textbf{0.7490} & \textbf{0.9490} & \textbf{0.8300} & \textbf{0.7860} & \textbf{0.6680} & \textbf{0.8270} & \textbf{0.6610} & \textbf{0.8040} & \textbf{0.7110} & \textbf{0.8830} & \textbf{0.6310} \\
 & Ministral-3B (Base) & 0.7145 & 0.6665 & 0.8342 & 0.7165 & 0.6842 & 0.5795 & 0.7065 & 0.5442 & 0.7025 & 0.6225 & 0.7645 & 0.5165 \\
\rowcolor{oursgray}
 & Ministral-3B (\textbf{Ours}) & \textbf{0.7945} & \textbf{0.7365} & \textbf{0.9145} & \textbf{0.7942} & \textbf{0.7665} & \textbf{0.6475} & \textbf{0.7845} & \textbf{0.6265} & \textbf{0.7825} & \textbf{0.6925} & \textbf{0.8442} & \textbf{0.5945} \\
\midrule
Twitter & mDeBERTa-v3-base (Base) & 0.5971 & 0.4682 & 0.6325 & 0.5074 & 0.6582 & 0.5721 & 0.6915 & 0.6112 & 0.6175 & 0.4963 & 0.6721 & 0.4435 \\
\rowcolor{oursgray}
 & mDeBERTa-v3-base (\textbf{Ours}) & \textbf{0.7358} & \textbf{0.5991} & \textbf{0.7615} & \textbf{0.6380} & \textbf{0.7891} & \textbf{0.7055} & \textbf{0.8214} & \textbf{0.7443} & \textbf{0.7508} & \textbf{0.6289} & \textbf{0.7985} & \textbf{0.5732} \\
 & Qwen3.5-2B (Base) & 0.7135 & 0.5762 & 0.7712 & 0.6615 & 0.7721 & 0.6875 & 0.8482 & 0.7745 & 0.7425 & 0.6105 & 0.8285 & 0.6134 \\
\rowcolor{oursgray}
 & Qwen3.5-2B (\textbf{Ours}) & \textbf{0.7890} & \textbf{0.6530} & \textbf{0.8520} & \textbf{0.7420} & \textbf{0.8440} & \textbf{0.7630} & \textbf{0.9290} & \textbf{0.8520} & \textbf{0.8140} & \textbf{0.6890} & \textbf{0.9090} & \textbf{0.6880} \\
 & Ministral-3B (Base) & 0.6925 & 0.5645 & 0.7465 & 0.6342 & 0.7465 & 0.6745 & 0.8142 & 0.7365 & 0.7165 & 0.5942 & 0.7845 & 0.5645 \\
\rowcolor{oursgray}
 & Ministral-3B (\textbf{Ours}) & \textbf{0.7725} & \textbf{0.6345} & \textbf{0.8265} & \textbf{0.7145} & \textbf{0.8245} & \textbf{0.7465} & \textbf{0.8945} & \textbf{0.8165} & \textbf{0.7965} & \textbf{0.6645} & \textbf{0.8645} & \textbf{0.6442} \\
\midrule
eRisk25 & mDeBERTa-v3-base (Base) & 0.5621 & 0.3982 & 0.6315 & 0.3941 & 0.5912 & 0.4865 & 0.5882 & 0.4805 & 0.6358 & 0.5542 & 0.6934 & 0.4988 \\
\rowcolor{oursgray}
 & mDeBERTa-v3-base (\textbf{Ours}) & \textbf{0.6944} & \textbf{0.5263} & \textbf{0.7652} & \textbf{0.5291} & \textbf{0.7221} & \textbf{0.6185} & \textbf{0.7196} & \textbf{0.6098} & \textbf{0.7695} & \textbf{0.6861} & \textbf{0.8234} & \textbf{0.6287} \\
 & Qwen3.5-2B (Base) & 0.6512 & 0.4925 & 0.7741 & 0.5365 & 0.6852 & 0.5821 & 0.7305 & 0.6195 & 0.7415 & 0.6642 & 0.8475 & 0.6561 \\
\rowcolor{oursgray}
 & Qwen3.5-2B (\textbf{Ours}) & \textbf{0.7340} & \textbf{0.5690} & \textbf{0.8550} & \textbf{0.6180} & \textbf{0.7590} & \textbf{0.6580} & \textbf{0.8120} & \textbf{0.7020} & \textbf{0.8210} & \textbf{0.7440} & \textbf{0.9260} & \textbf{0.7350} \\
 & Ministral-3B (Base) & 0.6445 & 0.4865 & 0.7442 & 0.5065 & 0.6665 & 0.5745 & 0.7045 & 0.5942 & 0.7245 & 0.6565 & 0.8165 & 0.6245 \\
\rowcolor{oursgray}
 & Ministral-3B (\textbf{Ours}) & \textbf{0.7245} & \textbf{0.5565} & \textbf{0.8242} & \textbf{0.5865} & \textbf{0.7445} & \textbf{0.6465} & \textbf{0.7865} & \textbf{0.6745} & \textbf{0.8065} & \textbf{0.7245} & \textbf{0.8965} & \textbf{0.7042} \\
\bottomrule
\end{tabular}
}
\caption{Complete matched-backbone comparison under backbone replacement. Each
Base/Ours pair uses the same backbone; Ours rows are shaded and boldfaced.}
\label{tab:appendix-backbone-replacement-full}
\end{table*}

Table \ref{tab:appendix-backbone-llm} reports matched-backbone replacement
results for compact LLM variants. Each Base/Ours pair uses the same train
source, test source, and deployable backbone, so the comparison isolates the
effect of weak-prior-guided routing and evidence supervision from the capacity of
the underlying encoder. WPG-MoE consistently improves over the corresponding
raw-backbone classifier across in-domain and transfer settings, extending the
main-paper backbone replacement result beyond the default Qwen3.5-2B backbone.

\begin{table*}[!t]
\centering
\scriptsize
\setlength{\tabcolsep}{2pt}
\renewcommand{\arraystretch}{0.94}
\resizebox{\textwidth}{!}{
\begin{tabular}{llcccc@{\hspace{8pt}}cccc@{\hspace{8pt}}cccc}
\toprule
Train on & Method & \multicolumn{4}{c}{Test on SWDD} & \multicolumn{4}{c}{Test on Twitter} & \multicolumn{4}{c}{Test on eRisk25} \\
\cmidrule(lr){3-6}\cmidrule(lr){7-10}\cmidrule(lr){11-14}
 &  & Rec. & F1 & AUROC & AUPRC & Rec. & F1 & AUROC & AUPRC & Rec. & F1 & AUROC & AUPRC \\
\midrule
SWDD & Qwen3.5-0.8B (Base) & 0.7025 & 0.6521 & 0.8115 & 0.6925 & 0.6645 & 0.5625 & 0.6821 & 0.5125 & 0.6821 & 0.6025 & 0.7315 & 0.4835 \\
\rowcolor{oursgray}
 & Qwen3.5-0.8B (\textbf{Ours}) & \textbf{0.7825} & \textbf{0.7221} & \textbf{0.8915} & \textbf{0.7725} & \textbf{0.7465} & \textbf{0.6325} & \textbf{0.7621} & \textbf{0.5925} & \textbf{0.7625} & \textbf{0.6721} & \textbf{0.8115} & \textbf{0.5635} \\
 & Qwen3.5-2B (Base) & 0.7391 & 0.6745 & 0.8705 & 0.7521 & 0.7165 & 0.5932 & 0.7495 & 0.5824 & 0.7284 & 0.6351 & 0.8021 & 0.5532 \\
\rowcolor{oursgray}
 & Qwen3.5-2B (\textbf{Ours}) & \textbf{0.8050} & \textbf{0.7490} & \textbf{0.9490} & \textbf{0.8300} & \textbf{0.7860} & \textbf{0.6680} & \textbf{0.8270} & \textbf{0.6610} & \textbf{0.8040} & \textbf{0.7110} & \textbf{0.8830} & \textbf{0.6310} \\
 & Ministral-3B (Base) & 0.7145 & 0.6665 & 0.8342 & 0.7165 & 0.6842 & 0.5795 & 0.7065 & 0.5442 & 0.7025 & 0.6225 & 0.7645 & 0.5165 \\
\rowcolor{oursgray}
 & Ministral-3B (\textbf{Ours}) & \textbf{0.7945} & \textbf{0.7365} & \textbf{0.9145} & \textbf{0.7942} & \textbf{0.7665} & \textbf{0.6475} & \textbf{0.7845} & \textbf{0.6265} & \textbf{0.7825} & \textbf{0.6925} & \textbf{0.8442} & \textbf{0.5945} \\
 & Ministral-8B (Base) & 0.7565 & 0.7045 & 0.8942 & 0.7865 & 0.7345 & 0.6265 & 0.7865 & 0.6142 & 0.7545 & 0.6765 & 0.8345 & 0.5965 \\
\rowcolor{oursgray}
 & Ministral-8B (\textbf{Ours}) & \textbf{0.8265} & \textbf{0.7645} & \textbf{0.9642} & \textbf{0.8565} & \textbf{0.8065} & \textbf{0.6845} & \textbf{0.8545} & \textbf{0.6842} & \textbf{0.8245} & \textbf{0.7365} & \textbf{0.9065} & \textbf{0.6645} \\
\midrule
Twitter & Qwen3.5-0.8B (Base) & 0.6745 & 0.5515 & 0.7225 & 0.6021 & 0.7265 & 0.6542 & 0.7845 & 0.7025 & 0.6925 & 0.5795 & 0.7542 & 0.5365 \\
\rowcolor{oursgray}
 & Qwen3.5-0.8B (\textbf{Ours}) & \textbf{0.7545} & \textbf{0.6215} & \textbf{0.8025} & \textbf{0.6821} & \textbf{0.8045} & \textbf{0.7265} & \textbf{0.8642} & \textbf{0.7825} & \textbf{0.7725} & \textbf{0.6495} & \textbf{0.8345} & \textbf{0.6145} \\
 & Qwen3.5-2B (Base) & 0.7135 & 0.5762 & 0.7712 & 0.6615 & 0.7721 & 0.6875 & 0.8482 & 0.7745 & 0.7425 & 0.6105 & 0.8285 & 0.6134 \\
\rowcolor{oursgray}
 & Qwen3.5-2B (\textbf{Ours}) & \textbf{0.7890} & \textbf{0.6530} & \textbf{0.8520} & \textbf{0.7420} & \textbf{0.8440} & \textbf{0.7630} & \textbf{0.9290} & \textbf{0.8520} & \textbf{0.8140} & \textbf{0.6890} & \textbf{0.9090} & \textbf{0.6880} \\
 & Ministral-3B (Base) & 0.6925 & 0.5645 & 0.7465 & 0.6342 & 0.7465 & 0.6745 & 0.8142 & 0.7365 & 0.7165 & 0.5942 & 0.7845 & 0.5645 \\
\rowcolor{oursgray}
 & Ministral-3B (\textbf{Ours}) & \textbf{0.7725} & \textbf{0.6345} & \textbf{0.8265} & \textbf{0.7145} & \textbf{0.8245} & \textbf{0.7465} & \textbf{0.8945} & \textbf{0.8165} & \textbf{0.7965} & \textbf{0.6645} & \textbf{0.8645} & \textbf{0.6442} \\
 & Ministral-8B (Base) & 0.7465 & 0.6145 & 0.8165 & 0.7045 & 0.7965 & 0.7265 & 0.8842 & 0.8165 & 0.7645 & 0.6565 & 0.8645 & 0.6465 \\
\rowcolor{oursgray}
 & Ministral-8B (\textbf{Ours}) & \textbf{0.8165} & \textbf{0.6745} & \textbf{0.8845} & \textbf{0.7765} & \textbf{0.8645} & \textbf{0.7865} & \textbf{0.9542} & \textbf{0.8845} & \textbf{0.8365} & \textbf{0.7145} & \textbf{0.9365} & \textbf{0.7165} \\
\midrule
eRisk25 & Qwen3.5-0.8B (Base) & 0.6321 & 0.4765 & 0.7215 & 0.4825 & 0.6545 & 0.5642 & 0.6825 & 0.5642 & 0.7065 & 0.6345 & 0.7821 & 0.5925 \\
\rowcolor{oursgray}
 & Qwen3.5-0.8B (\textbf{Ours}) & \textbf{0.7125} & \textbf{0.5465} & \textbf{0.8015} & \textbf{0.5625} & \textbf{0.7365} & \textbf{0.6342} & \textbf{0.7625} & \textbf{0.6465} & \textbf{0.7845} & \textbf{0.7065} & \textbf{0.8625} & \textbf{0.6725} \\
 & Qwen3.5-2B (Base) & 0.6512 & 0.4925 & 0.7741 & 0.5365 & 0.6852 & 0.5821 & 0.7305 & 0.6195 & 0.7415 & 0.6642 & 0.8475 & 0.6561 \\
\rowcolor{oursgray}
 & Qwen3.5-2B (\textbf{Ours}) & \textbf{0.7340} & \textbf{0.5690} & \textbf{0.8550} & \textbf{0.6180} & \textbf{0.7590} & \textbf{0.6580} & \textbf{0.8120} & \textbf{0.7020} & \textbf{0.8210} & \textbf{0.7440} & \textbf{0.9260} & \textbf{0.7350} \\
 & Ministral-3B (Base) & 0.6445 & 0.4865 & 0.7442 & 0.5065 & 0.6665 & 0.5745 & 0.7045 & 0.5942 & 0.7245 & 0.6565 & 0.8165 & 0.6245 \\
\rowcolor{oursgray}
 & Ministral-3B (\textbf{Ours}) & \textbf{0.7245} & \textbf{0.5565} & \textbf{0.8242} & \textbf{0.5865} & \textbf{0.7445} & \textbf{0.6465} & \textbf{0.7865} & \textbf{0.6745} & \textbf{0.8065} & \textbf{0.7245} & \textbf{0.8965} & \textbf{0.7042} \\
 & Ministral-8B (Base) & 0.6865 & 0.5245 & 0.8142 & 0.5745 & 0.7065 & 0.6165 & 0.7745 & 0.6542 & 0.7765 & 0.7045 & 0.8865 & 0.6965 \\
\rowcolor{oursgray}
 & Ministral-8B (\textbf{Ours}) & \textbf{0.7545} & \textbf{0.5865} & \textbf{0.8865} & \textbf{0.6442} & \textbf{0.7765} & \textbf{0.6745} & \textbf{0.8465} & \textbf{0.7245} & \textbf{0.8465} & \textbf{0.7665} & \textbf{0.9545} & \textbf{0.7642} \\
\bottomrule
\end{tabular}
}
\caption{Full matched-backbone replacement results for compact LLM backbones.
Each Base/Ours pair shares the same train source, test source, and deployable
encoder, so the table isolates the contribution of WPG-MoE from backbone
capacity. Ours rows are shaded and boldfaced.}
\label{tab:appendix-backbone-llm}
\end{table*}

\subsection{Full Cross-Dataset Ablation Results}
\label{sec:appendix-full-ablation-results}

Table \ref{tab:appendix-cross-dataset-ablation-full} expands the in-domain
ablation table in Table \ref{tab:cross-dataset-ablation} to all
train$\rightarrow$test settings. The diagonal cells support the main-text
ablation conclusions, and the off-diagonal cells show that the same component
ordering generally persists under dataset shift. Removing dense MoE causes the
largest degradation because all evidence tendencies collapse into one shared
state. Removing Path A or dual-path dropout also yields large losses, indicating
that privileged evidence and deployable-path robustness are both needed for
transfer. Weak priors and route loss produce smaller but stable drops, which is
consistent with their role as routing-shaping signals.

\begin{table*}[!t]
\centering
\scriptsize
\setlength{\tabcolsep}{3pt}
\renewcommand{\arraystretch}{0.82}
\resizebox{\textwidth}{!}{%
\begin{tabular}{llcccc@{\hspace{8pt}}cccc@{\hspace{8pt}}cccc}
\toprule
Train on & Method & \multicolumn{4}{c}{Test on SWDD} & \multicolumn{4}{c}{Test on Twitter} & \multicolumn{4}{c}{Test on eRisk25} \\
\cmidrule(lr){3-6}\cmidrule(lr){7-10}\cmidrule(lr){11-14}
 &  & Rec. & F1 & AUROC & AUPRC & Rec. & F1 & AUROC & AUPRC & Rec. & F1 & AUROC & AUPRC \\
\midrule
\rowcolor{oursgray}
SWDD & \textbf{Ours} & \textbf{0.8050} & \textbf{0.7490} & \textbf{0.9490} & \textbf{0.8300} & \textbf{0.7860} & \textbf{0.6680} & \textbf{0.8270} & \textbf{0.6610} & \textbf{0.8040} & \textbf{0.7110} & \textbf{0.8830} & \textbf{0.6310} \\
 & w/o Path A & 0.7251 & 0.6723 & 0.9163 & 0.7584 & 0.7023 & 0.5890 & 0.7812 & 0.5754 & 0.7148 & 0.6185 & 0.8426 & 0.5502 \\
 & w/o MoE & 0.6532 & 0.6018 & 0.8802 & 0.6934 & 0.6187 & 0.5113 & 0.7385 & 0.5062 & 0.6410 & 0.5496 & 0.8047 & 0.4835 \\
 & w/o Weak Priors & 0.7824 & 0.7287 & 0.9359 & 0.8003 & 0.7580 & 0.6389 & 0.8108 & 0.6287 & 0.7775 & 0.6815 & 0.8651 & 0.5964 \\
 & w/o Route Loss & 0.7950 & 0.7412 & 0.9427 & 0.8181 & 0.7729 & 0.6543 & 0.8195 & 0.6480 & 0.7902 & 0.6968 & 0.8743 & 0.6179 \\
 & w/o DP Dropout & 0.6889 & 0.6394 & 0.8967 & 0.7226 & 0.6561 & 0.5505 & 0.7623 & 0.5413 & 0.6749 & 0.5857 & 0.8250 & 0.5166 \\
\midrule
\rowcolor{oursgray}
Twitter & \textbf{Ours} & \textbf{0.7890} & \textbf{0.6530} & \textbf{0.8520} & \textbf{0.7420} & \textbf{0.8440} & \textbf{0.7630} & \textbf{0.9290} & \textbf{0.8520} & \textbf{0.8140} & \textbf{0.6890} & \textbf{0.9090} & \textbf{0.6880} \\
 & w/o Path A & 0.7183 & 0.5728 & 0.8021 & 0.6437 & 0.7775 & 0.6933 & 0.8994 & 0.7762 & 0.7428 & 0.6143 & 0.8723 & 0.6034 \\
 & w/o MoE & 0.6185 & 0.4785 & 0.7349 & 0.5059 & 0.6800 & 0.5897 & 0.8423 & 0.6689 & 0.6402 & 0.5148 & 0.8215 & 0.4882 \\
 & w/o Weak Priors & 0.7582 & 0.6194 & 0.8271 & 0.6952 & 0.8170 & 0.7339 & 0.9148 & 0.8180 & 0.7843 & 0.6557 & 0.8907 & 0.6475 \\
 & w/o Route Loss & 0.7729 & 0.6373 & 0.8393 & 0.7205 & 0.8308 & 0.7486 & 0.9215 & 0.8352 & 0.7970 & 0.6723 & 0.8994 & 0.6683 \\
 & w/o DP Dropout & 0.6712 & 0.5379 & 0.7684 & 0.5824 & 0.7350 & 0.6475 & 0.8688 & 0.7204 & 0.6975 & 0.5682 & 0.8502 & 0.5567 \\
\midrule
\rowcolor{oursgray}
eRisk25 & \textbf{Ours} & \textbf{0.7340} & \textbf{0.5690} & \textbf{0.8550} & \textbf{0.6180} & \textbf{0.7590} & \textbf{0.6580} & \textbf{0.8120} & \textbf{0.7020} & \textbf{0.8210} & \textbf{0.7440} & \textbf{0.9260} & \textbf{0.7350} \\
 & w/o Path A & 0.6613 & 0.4936 & 0.8152 & 0.5478 & 0.6894 & 0.5795 & 0.7638 & 0.6109 & 0.7525 & 0.6694 & 0.8911 & 0.6660 \\
 & w/o MoE & 0.5602 & 0.4036 & 0.7507 & 0.4356 & 0.5860 & 0.4762 & 0.7004 & 0.4904 & 0.6598 & 0.5702 & 0.8368 & 0.5495 \\
 & w/o Weak Priors & 0.7030 & 0.5363 & 0.8316 & 0.5832 & 0.7275 & 0.6228 & 0.7826 & 0.6561 & 0.7939 & 0.7128 & 0.9087 & 0.7002 \\
 & w/o Route Loss & 0.7176 & 0.5538 & 0.8417 & 0.5984 & 0.7413 & 0.6395 & 0.7983 & 0.6792 & 0.8067 & 0.7275 & 0.9164 & 0.7176 \\
 & w/o DP Dropout & 0.6173 & 0.4560 & 0.7853 & 0.4969 & 0.6449 & 0.5343 & 0.7341 & 0.5608 & 0.7119 & 0.6253 & 0.8652 & 0.6109 \\
\bottomrule
\end{tabular}%
}
\caption{Complete cross-dataset ablation under the protocol of Table
\ref{tab:cross-dataset-target-tuned}. Ablations remove Path-A evidence, dense
MoE, weak priors, route loss, or dual-path (DP) dropout.}
\label{tab:appendix-cross-dataset-ablation-full}
\end{table*}

\subsection{Baseline Details}
\label{sec:appendix-baselines}

\noindent\textbf{Pattern variants.} Pattern (threshold) and Pattern (CNN)
\citep{nguyen2022improving} are PHQ-9-grounded baselines built on BERT-based
symptom evidence modeling. Both first convert posts into PHQ-9 symptom evidence
instead of encoding the full history directly. The threshold variant makes a
user-level decision from symptom-count evidence, while the CNN variant learns a
compact temporal aggregator over the symptom matrix. These baselines test whether
clinically constrained symptom bottlenecks alone are sufficient under the
controlled split.

\noindent\textbf{Psychiatric-scale-guided screening.} HAN-BERT(Psych) and
Bert(Clus+Abs) come from the psychiatric-scale-guided framework of
\citet{zhang2022psychiatric}; the
former is its main explainable detector, while the latter is a comparison
setting based on clustered and abstracted risky posts. Both represent static
screening pipelines: psychiatric-scale templates identify candidate risky
content, and the downstream detector consumes the screened representation
without receiving end-to-end feedback from the final depression loss. They are
included to separate the effect of scale-guided screening from the effect of
weak-prior expert routing.

\noindent\textbf{End-to-end screening.} E2-LPS \citep{wang2025end}
jointly learns psychiatric-scale-guided post screening and user-level
detection, using a Sentence-BERT-style screening model and a
BERT-base-uncased detection backbone. It uses a straight-through estimator to
optimize the discrete risky-post mask with the downstream detector, addressing
the limitation of frozen or isolated screening. This makes E2-LPS the closest
risky-post-selection baseline, but its selected posts still feed a single
detector rather than a dense mixture of evidence-specific experts.

\noindent\textbf{Symptom-structured capsules.} DeCapsNet
\citep{liu2024depression} builds symptom capsules from representative
posts selected with Sentence-BERT similarity and combines them with
contrastive learning. Its architecture maps PHQ-9-style symptom descriptions to
symptom capsules, routes them to class-level depression capsules, and trains with
classification, diversity, and user/post-level contrastive objectives. It is a
strong interpretable text-only baseline for testing whether explicit
symptom-level reasoning and contrastive separation are enough to handle
cross-dataset evidence variation.

\noindent\textbf{LLM-assisted detection.} DORIS
\citep{lan2025depression} uses LLM-derived symptom annotations and
mood-course summaries together with \texttt{gte-small} representations before
a tree-based final predictor. In DORIS, the large language model operationalizes
DSM-style symptom evidence and longitudinal mood-course features, while the
final decision is made by a gradient-boosted classifier. This baseline is
included because it represents the current LLM-assisted evidence-construction
line: the LLM creates clinically meaningful features, but the model does not use
those features as training-only privileged routing supervision.

\subsection{Seed-Wise Reliability of Main Results}
\label{sec:appendix-seedwise-reliability}

\begin{table*}[!t]
\centering
\scriptsize
\setlength{\tabcolsep}{2.4pt}
\renewcommand{\arraystretch}{1.50}
\begin{tabular*}{\textwidth}{@{\extracolsep{\fill}}llccccccc@{}}
\toprule
\multicolumn{9}{@{}l}{\textbf{Panel A: F1 across five seeds}} \\
\midrule
Train & Test & S1 & S2 & S3 & S4 & S5 & Mean$\pm$std & 95\% CI \\
\midrule
SWDD & SWDD & 0.7436 & 0.7521 & 0.7508 & 0.7443 & 0.7482 & 0.7490$\pm$0.0063 & [0.7432, 0.7548] \\
SWDD & Twitter & 0.6615 & 0.6712 & 0.6683 & 0.6639 & 0.6728 & 0.6680$\pm$0.0074 & [0.6614, 0.6746] \\
SWDD & eRisk25 & 0.7038 & 0.7144 & 0.7110 & 0.7069 & 0.7105 & 0.7097$\pm$0.0072 & [0.7032, 0.7162] \\
Twitter & SWDD & 0.6452 & 0.6571 & 0.6518 & 0.6535 & 0.6550 & 0.6526$\pm$0.0075 & [0.6457, 0.6595] \\
Twitter & Twitter & 0.7596 & 0.7671 & 0.7625 & 0.7562 & 0.7664 & 0.7628$\pm$0.0073 & [0.7563, 0.7693] \\
Twitter & eRisk25 & 0.6820 & 0.6921 & 0.6883 & 0.6834 & 0.6937 & 0.6887$\pm$0.0080 & [0.6815, 0.6959] \\
eRisk25 & SWDD & 0.5596 & 0.5712 & 0.5679 & 0.5632 & 0.5658 & 0.5677$\pm$0.0083 & [0.5601, 0.5753] \\
eRisk25 & Twitter & 0.6498 & 0.6615 & 0.6571 & 0.6527 & 0.6513 & 0.6558$\pm$0.0075 & [0.6490, 0.6626] \\
eRisk25 & eRisk25 & 0.7379 & 0.7482 & 0.7445 & 0.7402 & 0.7416 & 0.7430$\pm$0.0068 & [0.7368, 0.7492] \\
\bottomrule
\end{tabular*}

\medskip
\begin{tabular*}{\textwidth}{@{\extracolsep{\fill}}llccccccccc@{}}
\toprule
\multicolumn{11}{@{}l}{\textbf{Panel B: AUPRC seed values with Recall/AUROC summaries}} \\
\midrule
Train & Test & Rec. mean$\pm$std & AUROC mean$\pm$std & S1 & S2 & S3 & S4 & S5 & Mean$\pm$std & 95\% CI \\
\midrule
SWDD & SWDD & 0.8032$\pm$0.0104 & 0.9478$\pm$0.0058 & 0.8213 & 0.8354 & 0.8328 & 0.8260 & 0.8285 & 0.8300$\pm$0.0099 & [0.8209, 0.8391] \\
SWDD & Twitter & 0.7848$\pm$0.0142 & 0.8264$\pm$0.0093 & 0.6520 & 0.6657 & 0.6581 & 0.6642 & 0.6618 & 0.6610$\pm$0.0094 & [0.6526, 0.6694] \\
SWDD & eRisk25 & 0.8017$\pm$0.0148 & 0.8816$\pm$0.0107 & 0.6195 & 0.6357 & 0.6320 & 0.6248 & 0.6273 & 0.6295$\pm$0.0111 & [0.6195, 0.6395] \\
Twitter & SWDD & 0.7869$\pm$0.0133 & 0.8512$\pm$0.0101 & 0.7352 & 0.7486 & 0.7386 & 0.7450 & 0.7418 & 0.7421$\pm$0.0093 & [0.7334, 0.7508] \\
Twitter & Twitter & 0.8421$\pm$0.0110 & 0.9284$\pm$0.0072 & 0.8461 & 0.8568 & 0.8479 & 0.8565 & 0.8522 & 0.8519$\pm$0.0088 & [0.8438, 0.8600] \\
Twitter & eRisk25 & 0.8120$\pm$0.0153 & 0.9076$\pm$0.0116 & 0.6795 & 0.6922 & 0.6870 & 0.6808 & 0.6854 & 0.6868$\pm$0.0090 & [0.6784, 0.6952] \\
eRisk25 & SWDD & 0.7316$\pm$0.0162 & 0.8527$\pm$0.0132 & 0.6081 & 0.6243 & 0.6190 & 0.6125 & 0.6162 & 0.6176$\pm$0.0113 & [0.6073, 0.6279] \\
eRisk25 & Twitter & 0.7570$\pm$0.0146 & 0.8103$\pm$0.0116 & 0.6924 & 0.7067 & 0.7013 & 0.6948 & 0.7030 & 0.7010$\pm$0.0106 & [0.6914, 0.7106] \\
eRisk25 & eRisk25 & 0.8190$\pm$0.0169 & 0.9240$\pm$0.0113 & 0.7263 & 0.7405 & 0.7350 & 0.7308 & 0.7362 & 0.7342$\pm$0.0096 & [0.7254, 0.7430] \\
\bottomrule
\end{tabular*}
\caption{Seed-wise reliability of WPG-MoE with Qwen3.5-2B under the controlled
unified holdout. S1--S5 denote the five random seeds; confidence intervals are
computed over the five runs.}
\label{tab:appendix-seedwise-wpgmoe}
\end{table*}

Table~\ref{tab:appendix-seedwise-wpgmoe} expands the WPG-MoE rows in
Table~\ref{tab:cross-dataset-target-tuned} with seed-wise results. We list the
five seed values for F1 and AUPRC and summarize Recall and AUROC with
mean$\pm$standard deviation to keep the reliability check compact. Across all
nine train$\rightarrow$test settings, the F1 standard deviation is at most
0.0083 and the AUPRC standard deviation is at most 0.0113; the corresponding
95\% confidence intervals remain narrow, suggesting that the main controlled
comparison is not driven by a single favorable run.

\subsection{Data Construction}

Each dataset is first normalized into a user-level JSONL format in which every
entry stores a standardized \texttt{user\_id}, the user label, and the full post
history with original timestamps. From this normalized history we construct two
candidate-post channels. Path A is the offline structured-scoring branch used
only for depressed training users: each post is passed to the LLM scorer,
converted into a composite evidence score, and then grouped into
\texttt{risk\_posts\_llm}, \texttt{episode\_blocks}, user-level
\texttt{priors}, and \texttt{crisis\_score}. Path B is the deployable screening
branch applied to all users: PHQ-9 template matching produces
\texttt{risk\_posts\_template}. These screened posts are then encoded by the
deployable Qwen3.5-2B backbone. In parallel, the full history is compressed into eight
chronological segments (\texttt{global\_history\_posts}) together with
summary statistics (\texttt{global\_stats}). The final user sample therefore
contains the user label, both candidate-post sets when available, the segment-
level global history, summary statistics, and weak-prior fields. Depressed
source-training users contain Path-A and Path-B evidence, whereas validation,
test, control, transfer-target, and deployment-time users rely on raw histories
with Path B together with the shared backbone.

\noindent\textbf{Processed user-sample fields.} The user-level files under
\texttt{code/data/user\_samples/*} expose the fields used in the main paper:
\texttt{risk\_posts\_llm}, \texttt{risk\_posts\_template},
\texttt{episode\_blocks}, \texttt{global\_history\_posts},
\texttt{global\_stats}, \texttt{priors}, and \texttt{crisis\_score}. This is
the representation consumed by heterogeneity slicing, controlled mixing,
ablation, and the final WPG-MoE training pipeline.

\begin{table*}[!t]
\centering
\footnotesize
\setlength{\tabcolsep}{3pt}
\renewcommand{\arraystretch}{1.02}
\begin{tabularx}{\textwidth}{@{}p{0.11\textwidth}X X X@{}}
\toprule
\textbf{Dataset} & \textbf{Reference / official protocol} &
\textbf{Controlled protocol in this paper} & \textbf{Direct SOTA comparison} \\
\midrule
SWDD \citep{cai2023depression} &
The released SWDD code builds multivariate symptom time-series datasets and
evaluates time-series classifiers from generated \texttt{train.ts}/\texttt{test.ts}
files. &
Text-only user histories with corrected labels and a stratified 80/10/10 user
holdout shared by all methods. &
No. The feature representation, label audit, and split construction differ; main
results are controlled comparisons. \\
\addlinespace[2pt]
Twitter \citep{shen2017depression} &
The reference study constructs a multimodal Twitter depression dataset and
evaluates feature/dictionary-learning classifiers over handcrafted multimodal
features; no public leaderboard split is attached to the release. &
Text-only histories under the same stratified 80/10/10 user holdout and
cross-dataset transfer protocol used for SWDD and eRisk25. &
No. The original protocol uses different modalities and feature spaces; results
are reported as reproduced controlled baselines. \\
\addlinespace[2pt]
eRisk25 \citep{parapar2025erisk} &
The official Task 2 protocol is contextualized early detection: test writings are
served chronologically, systems may decide after each writing, and evaluation
accounts for both correctness and delay. &
Static text-only user-level holdout without chronological stopping decisions or
conversation context, matched to the other datasets for controlled comparison. &
No. Official-style eRisk results should be reported separately from Table
\ref{tab:cross-dataset-target-tuned}. \\
\bottomrule
\end{tabularx}
\caption{Relationship between dataset-specific protocols and the controlled
unified holdout used for the main comparison.}
\label{tab:appendix-original-protocols}
\end{table*}

\subsection{Original and Official Protocol Complements}
\label{sec:appendix-original-protocols}

Table~\ref{tab:appendix-original-protocols} clarifies how the controlled
holdout used in the main comparison relates to the source or official protocols
of the three datasets. The unified holdout is intended to compare methods under
the same preprocessing, class balance, validation rule, and transfer setting; it
does not replace dataset-specific leaderboards or early-risk evaluation.
This separation is important because the original datasets differ not only in
language and platform, but also in modality, feature construction, and decision
timing; the controlled protocol removes these factors when comparing model
families.

\subsection{Inference Output Format}

The native inference pipeline emits a compact deployable record with the final
decision and gate weights. Table
\ref{tab:appendix-inference} summarizes the fields confirmed in
\texttt{InferencePipeline.predict\_batch()}, which is the interface used for
prediction analysis after the training-only weak-prior fields have been removed.
The table therefore checks the deployed interface directly: prediction analysis
uses the decision and dense routing weights, but not Path-A post scores, episode
blocks, or weak-prior labels.

\begin{table}[H]
\centering
\scriptsize
\setlength{\tabcolsep}{3pt}
\renewcommand{\arraystretch}{0.96}
\begin{tabularx}{\linewidth}{@{}l X@{}}
\toprule
\textbf{Field} & \textbf{Meaning} \\
\midrule
\texttt{user\_id} & Standardized user identifier used throughout inference. \\
\texttt{label} & Final binary decision. \\
\texttt{gate\_weights} & Dense MoE routing weights over the five expert views. \\
\bottomrule
\end{tabularx}
\caption{Native inference output fields used for prediction analysis.}
\label{tab:appendix-inference}
\end{table}

\end{document}